\RequirePackage{fix-cm}
\documentclass[smallextended]{svjour3}       
\smartqed  
\usepackage{graphicx}
\usepackage{graphicx}
\usepackage{bm}
\usepackage{mathrsfs}
\usepackage{amsmath}
\usepackage{amssymb}
\usepackage{algorithm}
\usepackage{algorithmic}
\usepackage{array}
\usepackage{bm}
\usepackage{booktabs}
\usepackage{caption3}
\usepackage{url}
\usepackage{multirow}
\usepackage{xcolor}
\usepackage{color}
\newcommand{\etal}{\textit{et al.~}}
\newcommand{\ie}{i.e.}

\begin{document}

\title{Multiplication fusion of sparse and collaborative-competitive representation for image classification
}


\author{Zi-Qi Li$^{1,2}$         \and
        Jun Sun$^{1,2*}$          \and
        Xiao-Jun Wu$^{1,2}$      \and
        He-Feng Yin$^{1,2}$ 
}


\institute{Jun Sun \at
	\email{junsun@jiangnan.edu.cn} 
	\\
	$^{1}$School of Internet of Things Engineering, Jiangnan University, Wuxi 214122, China
	\and
	$^{2}$Jiangsu Provincial Engineering Laboratory of Pattern Recognition and Computational Intelligence, Jiangnan University, Wuxi 214122, China
}

\date{Received: date / Accepted: date}

\maketitle

\begin{abstract}
Representation based classification methods have become a hot research topic during the past few years, and the two most prominent approaches are sparse representation based classification (SRC) and collaborative representation based classification (CRC). CRC reveals that it is the collaborative representation {\color{red}rather than} the sparsity that makes SRC successful. Nevertheless, the dense representation of CRC may not be discriminative which will degrade its performance for classification tasks. To {\color{red}alleviate} this problem to some extent, we propose a new method called sparse and collaborative-competitive representation based classification (SCCRC) for image classification. Firstly, the coefficients of the test sample are obtained by SRC and CCRC, respectively. Then the fused coefficient is derived by multiplying the coefficients of SRC and CCRC. Finally, the test sample is designated to the class that has the minimum residual. Experimental results on several benchmark databases demonstrate the efficacy of our proposed SCCRC. {\color{red}The source code of SCCRC is accessible at https://github.com/li-zi-qi/SCCRC.}
\end{abstract}

\section{Introduction}
\label{intro}
Representation based classification methods (RBCM) have already gained increasing attention in various research fields, e.g. character recognition~\cite{qu2018air}, person re-identification~\cite{prates2019kernel} and hyperspectral image classification~\cite{yang2018hyperspectral}. SRC~\cite{wright2009robust} is a pioneering work of RBCM, which directly uses all the training data as the dictionary to represent the test sample, and classifies the test sample by checking which class leads to the minimal reconstruction error. SRC solves an $\ell_1$-norm optimization problem, and thus when the size of dictionary is huge, the sparse decomposition process may be very slow. One way to speed up the sparse coding process is to reduce the size of dictionary by selecting representative training samples. Li \etal \cite{li2010local} proposed a local sparse representation based classification (LSRC) scheme, which performs sparse decomposition in local neighborhood. Similarly, Zhang \etal \cite{zhang2010k} presented KNN-SRC, which chooses $K$ nearest neighbors of a testing sample from all the training samples to represent the testing sample. Ortiz \etal \cite{ortiz2014face} developed a linearly approximated sparse representation-based classification (LASRC) algorithm that employs linear regression to perform sample selection for $\ell_1$-minimization. The other way is to obtain a compact and discriminative dictionary through dictionary learning. The most classic dictionary learning approach is KSVD~\cite{aharon2006k}, which is an iterative method that alternates between sparse coding and a process of updating the dictionary atoms. To make KSVD more suitable for classification tasks, Zhang \etal \cite{zhang2010discriminative} proposed a discriminative K-SVD (D-KSVD) method which incorporates the classification error into the objective function of KSVD. Jiang \etal \cite{jiang2013label} presented a label consistent K-SVD (LC-KSVD) algorithm which combines a new label consistency constraint with the reconstruction error and the classification error to form a unified objective function. Later Kviatkovsky \etal \cite{kviatkovsky2017equivalence} proved that under identical initialization conditions, LC-KSVD with uniform atom allocation is essentially a reformulation of DKSVD. Very recently, Song \etal \cite{song2018euler} designed a kernel dictionary learning approach called Euler Label Consistent K-SVD (ELC-KSVD) to capture the nonlinear similarity of features.

Another prominent approach of RBCM is CRC~\cite{zhang2011sparse} which replaces the $\ell_1$-norm in SRC with $\ell_2$-norm. Furthermore, Zhang \etal \cite{zhang2011sparse} revealed that it is the collaborative representation (CR) mechanism, but not the $\ell_1$-norm sparsity that makes SRC successful for classification. Likewise, Xu \etal \cite{xu2017new} introduced a discriminative sparse representation (DSR) method for robust face recognition via $\ell_2$ regularization. The representation fidelity term in DSR is measured by $\ell_2$-norm. To enhance the robustness of DSR, Zeng \etal \cite{zeng2017antinoise} proposed a robust version of DSR called Antinoise sparse representation method based on joint $\ell_1$ and $\ell_2$ regularization (Anti-L1L2) by employing the $\ell_1$-norm to measure the fidelity term. Although CRC presents a geometric interpretation of its working mechanism, it is hard to understand. Afterwards, Cai \etal \cite{cai2016probabilistic} analyzed the classification mechanism of CRC from a probabilistic viewpoint and proposed a probabilistic collaborative representation based classifier (ProCRC). Based on ProCRC, Lan \etal \cite{lan2018prior} explored a method called prior knowledge-based probabilistic CRC (PKPCRC) with further consideration of the prior knowledge extracted from the training samples. Yuan \etal \cite{yuan2018collaborative} constructed a collaborative-competitive representation based classifier (CCRC) model by introducing a competitive regularization term into the objective function of CRC. Using the training samples of all classes to collaboratively represent a test sample may produce negative effect, to overcome this problem, Zheng \etal \cite{zheng2019collaborative} presented a $k$-nearest classes based classification scheme. Moreover, Waqas \etal \cite{waqas2013collaborative} proposed a method known as collaborative neighbor representation classifier (CNRC) which represents a test sample over the whole training dictionary by automatically selecting bases from the training samples close to test sample. By integrating CNRC and DSR, Gou \etal \cite{gou2018new} designed a new discriminative collaborative neighbor representation (DCNR) method for face recognition.

The above RBCM and their variants emphasize too much on the role of $\ell_1$-norm sparsity or collaborative representation, and thus researchers are seeking to find new ways of combining them to enhance the performance of classification. Akhtar \etal \cite{akhtar2017efficient} argued that sparseness of collaborative representation explicitly contributes to accurate classification, and they developed a sparsity augmented collaborative representation based classification (SA-CRC) scheme. Zeng \etal \cite{zeng2017multiplication} proposed a representation-based image classification method that integrates SRC with CRC by a multiplication operation on the solutions. Li \etal \cite{li2016hyperspectral} presented a fused representation-based classification (FRC) method which attempts to achieve the balance between CR and SR in the residual domain. However, there is a weight parameter in FRC which needs to be set manually. {\color{red}Furthermore, the results in ~\cite{akhtar2017efficient} demonstrate that sparsity of collaborative representation does play a critical role in the correct classification of test samples. By multiplying the coefficient vectors obtained by SRC and CRC, sparsity of the fused coefficient vector in SCRC~\cite{zeng2017multiplication} is enhanced to some extent. Unfortunately, the coefficient vector obtained by CRC is limited sparse. In order to further promote the sparsity of fused coefficient vector,} in this paper, we propose a new method coined sparse and collaborative-competitive representation based classification (SCCRC) for image classification. The fused coefficient is obtained by multiplying the coefficients of SRC and CCRC, and then the test sample is classified by checking which class yields the least residual. {\color{red}Some representative RBCM and our proposed SCCRC are summarized in Table~\ref{tab:summary}, and our main contributions are summarized as follows,
\begin{enumerate}

\item Our proposed SCCRC involves the multiplication of coefficients obtained by SRC and CCRC,  it does not need to determine additional parameters, which makes SCCRC very efficient.

\item Although SCCRC is simple in principle, it outperforms some state-of-the-art RBCM on both clean and corrupted images in terms of classification accuracy.

\item Statistical significance test indicates that the performance differences between SCCRC and all the competing approaches are statistically significant.
\end{enumerate}}

\begin{table}[]
\caption{{\color{red}Summarization of some representative RBCM. Note that in FRC, the residuals obtained by SRC and CRC are fused by a weghting parameter $\theta$. In SCRC and SA-CRC, $\hat{\boldsymbol{\alpha}}$ and $\check{\boldsymbol{\alpha}}$ are the coefficient vectors obtained by SRC and CRC, respectively. In our proposed SCCRC, $\boldsymbol{\alpha}$ and $\boldsymbol{\beta}$ are the coefficient vectors obtained by SRC and CCRC, respectively.}}
\label{tab:summary}
\centering
\begin{tabular}{cc}
\hline
{\color{red}Algorithm} & {\color{red}Formulation}                     \\ \hline
{\color{red}SRC~\cite{wright2009robust}}      &   {\color{red}$\underset{\boldsymbol{\alpha}}{\min}\left\{ \left\| \boldsymbol{y}-\mathbf{X}\boldsymbol{\alpha} \right\|_{2}^{2}+\lambda\left\| \boldsymbol{\alpha} \right\|_{1} \right\}$}    \\
{\color{red}CRC~\cite{zhang2011sparse}}      & {\color{red}$  \underset{\boldsymbol{\alpha}}{\min}\left\{ \left\| \boldsymbol{y}-\mathbf{X}\boldsymbol{\alpha} \right\|_{2}^{2}+\lambda\left\| \boldsymbol{\alpha} \right\|_{2}^{2} \right\}$}         \\
{\color{red}FRC~\cite{li2016hyperspectral}}      & {\color{red}$\boldsymbol{r}=(1-\theta)\left\| \boldsymbol{y}-\mathbf{X}_{i}\boldsymbol{\hat{\alpha}_{i}} \right\|_{2}+\theta \left\| \boldsymbol{y}-\mathbf{X}_{i}\boldsymbol{\check{\alpha}_{i}} \right\|_{2}$}    \\
{\color{red}SCRC~\cite{zeng2017multiplication}}    & {\color{red}$\boldsymbol{f}=\hat{\boldsymbol{\alpha}} \odot \check{\boldsymbol{\alpha}}$}   \\
{\color{red}NRC~\cite{xu2019sparse}}      &  {\color{red}$\underset{\boldsymbol{\alpha}}{\min}\left\| \boldsymbol{y}-\mathbf{X}\boldsymbol{\alpha} \right\|_{2}^{2}, \ \textrm{s.t.} \ \boldsymbol{\alpha}\geq 0$}         \\
{\color{red}SA-CRC~\cite{akhtar2017efficient}}   & {\color{red}$\stackrel{\circ}{\boldsymbol{\alpha}}=\frac{\hat{\boldsymbol{\alpha}}+\check{\boldsymbol{\alpha}}}{\left\|\hat{\boldsymbol{\alpha}}+\check{\boldsymbol{\alpha}}\right\|_{2}}$}  \\
{\color{red}CCRC~\cite{yuan2018collaborative}}     & {\color{red}$\underset{\boldsymbol{\beta}}{\textrm{min}}\left\{\|\boldsymbol{y}-\mathbf{X} \boldsymbol{\beta}\|_{2}^{2}+\lambda_{1}\|\boldsymbol{\beta}\|_{2}^{2}+\lambda_{2} \sum_{i=1}^{C}\left\|\boldsymbol{y}-\mathbf{X}_{i} \boldsymbol{\beta}_{i}\right\|_{2}^{2}\right\}$}   \\
{\color{red}SCCRC}   &   {\color{red}$\boldsymbol{f}=\boldsymbol{\alpha} \odot \boldsymbol{\beta}$}    \\ \hline
\end{tabular}
\end{table}

The remainder of this paper is structured as follows: Section \ref{sec:sect_2} introduces several related approaches which include SRC, CRC, LRC~\cite{naseem2010linear} and CCRC. Section \ref{sec:sect_3} presents our SCCRC algorithm. Section \ref{sec:sect_4} reports the experiments on several benchmark databases. Finally, Section \ref{sec:sect_5} concludes the paper.

\section{Related work}
\label{sec:sect_2}
As our SCCRC is based on SRC and CCRC, we now briefly review these approaches for the sake of completeness.

We consider a set of $n$ training samples collected from $C$ subjects, each training image is represented as a vector corresponding to the $i$th column of the dictionary. Thus, all training samples form the matrix $\mathbf{X}=\left[\mathbf{X}_1,\mathbf{X}_2,\ldots,\mathbf{X}_C\right]\in\mathbb{R}^{d\times n}$ and $\mathbf{X}_i=\left[\mathbf{X}_{i,1},\mathbf{X}_{i,2},\ldots,\mathbf{X}_{i,n_i}\right]\in\mathbb{R}^{d\times n_i},i=1,2,\ldots,C$, where $d$ is the dimensionality of each sample and $n_i$ denotes the number of training samples in the $i$th class.

\subsection{Sparse representation based classification}
\label{sec:2}
Given a test sample $\boldsymbol{y}\in\mathbb{R}^{d\times 1}$, SRC employs a sparse linear superposition of all the training data to represent the test sample by solving the following $\ell_1$-norm minimization problem,

\begin{equation}\label{equ1}
  \underset{\boldsymbol{\alpha}}{\min}\left\{ \left\| \boldsymbol{y}-\mathbf{X}\boldsymbol{\alpha} \right\|_{2}^{2}+\lambda\left\| \boldsymbol{\alpha} \right\|_{1} \right\}
\end{equation}
where $\lambda>0$ is a balancing parameter. Then the reconstruction error (residual) for each class is obtained by,

\begin{equation}\label{equ2}
\boldsymbol{r}_{i}=\left\| \boldsymbol{y}-\mathbf{X}_{i}\boldsymbol{\hat{\alpha}_{i}} \right\|_{2}
\end{equation}
where $\boldsymbol{\hat{\alpha}_{i}}$ is the coefficients that correspond to the $i$th class. Finally, the identity of the test sample $\boldsymbol{y}$ is determined by evaluating which class leads to the least residual, i.e.

\begin{equation}\label{equ3}
\textrm{identity}\left( \boldsymbol{y} \right)=\underset{i}{\min}~\boldsymbol{r}_{i}
\end{equation}

\subsection{Collaborative representation based classification}
\label{sec:3}
CRC estimates the representation of the test sample $\boldsymbol{y}$ by relaxing the $\ell_1$-norm to the $\ell_2$-norm in (\ref{equ1}). The objective function of CRC is formulated as follows,

\begin{equation}\label{equ4}
  \underset{\boldsymbol{\alpha}}{\min}\left\{ \left\| \boldsymbol{y}-\mathbf{X}\boldsymbol{\alpha} \right\|_{2}^{2}+\lambda\left\| \boldsymbol{\alpha} \right\|_{2}^{2} \right\}
\end{equation}
where $\lambda>0$ is a balancing parameter. Eq. (\ref{equ4}) has closed form solution $\hat{\boldsymbol{\alpha}}=\left( \mathbf{X}^{T}\mathbf{X}+\lambda\mathbf{I} \right)^{-1}\mathbf{X}^{T}\boldsymbol{y}$. In~\cite{zhang2011sparse}, Zhang found that in addition to the residual, the classwise coefficients $\boldsymbol{\hat{\alpha}_{i}}$ can also bring some discrimination information for classification. Therefore, they proposed the following regularized residual for classification,

\begin{equation}\label{equ5}
\boldsymbol{r}_{i}=\frac{\left\| \boldsymbol{y}-\mathbf{X}_{i}{\boldsymbol{\hat{\alpha}}_{i}} \right\|_{2}}{\left\| {\boldsymbol{\hat{\alpha}}}_{i} \right\|_{2}}
\end{equation}

\subsection{Linear regression classification}
\label{sec:4}
Different from SRC and CRC which employ all the training samples for representation, the mechanism of LRC adopts the training samples in each class to reconstruct the test sample $\boldsymbol{y}$ in a classwise manner. Specifically, the objective function of LRC is formulated as,

\begin{equation}\label{equ6}
  \underset{\boldsymbol{\alpha}_{i}}{\min}\left\| \boldsymbol{y}-\mathbf{X}_{i}\boldsymbol{\alpha}_{i} \right\|_{2}^{2}
\end{equation}

Eq.~(\ref{equ6}) has closed-form solution, which is given by,

\begin{equation}\label{equ7}
{\boldsymbol{\hat{\alpha}}_{i}}=\left( {\mathbf{X}}_{i}^{T}{\mathbf{X}}_{i} \right)^{-1}\mathbf{X}_{i}^{T}\boldsymbol{y}
\end{equation}

Finally, the test sample $\boldsymbol{y}$ is classified according to the following rule,

\begin{equation}\label{equ8}
\textrm{identity}(\boldsymbol{y})=\underset{i}{\min}\left\| \boldsymbol{y}-{\mathbf{X}}_{i}{\boldsymbol{\hat{\alpha}}_{i}} \right\|_{2}
\end{equation}

In fact, the essence of LRC is nearest subspace (NS) classifier with downsampled features.

\subsection{Collaborative-competitive representation based classification (CCRC)}
\label{sec:5}
Yuan \etal \cite{yuan2018collaborative} proposed a collaborative-competitive representation based classifier model (CCRC) which incorporates a competitive term into the formulation of CRC, and the objective function of CCRC is formulated as follows,

\begin{equation}\label{equ9}
    \min _{\boldsymbol{\beta}}\left\{\|\boldsymbol{y}-\mathbf{X} \boldsymbol{\beta}\|_{2}^{2}+\lambda_{1}\|\boldsymbol{\beta}\|_{2}^{2}+\lambda_{2} \sum_{i=1}^{C}\left\|\boldsymbol{y}-\mathbf{X}_{i} \boldsymbol{\beta}_{i}\right\|_{2}^{2}\right\}
\end{equation}
where $\|\boldsymbol{y}-\mathbf{X} \boldsymbol{\beta}\|_{2}^{2}$ aims to collaboratively express the test sample by using all the training samples, $\sum_{i=1}^{C}\left\|\boldsymbol{y}-\mathbf{X}_{i} \boldsymbol{\beta}_{i}\right\|_{2}^{2}$ encourages the competitive representation across different classes, $\lambda_{1}$ and $\lambda_{2}$ are balancing parameters. As we can see from Eq.~(\ref{equ9}), if $\lambda_{2}$ equals to 0, CCRC boils down to CRC. CCRC has closed-form solution, which is given by,

\begin{equation}\label{equ10}
    \boldsymbol{\beta}=\mathbf{P} \boldsymbol{y}
\end{equation}
where $\mathbf{P}=\left(1+\lambda_{2}\right)\left(\mathbf{X}^{T} \mathbf{X}+\lambda_{1} \mathbf{I}+\lambda_{2} \mathbf{M}\right)^{-1} \mathbf{X}^{T}$ and $\mathbf{M}=\left[ \begin{array}{ccc}{\mathbf{X}_{1}^{T} \mathbf{X}_{1}} & {\cdots} & {0} \\ {\vdots} & {\ddots} & {\vdots} \\ {0} & {\cdots} & {\mathbf{X}_{C}^{T} \mathbf{X}_{C}}\end{array}\right]$

\section{Sparse and collaborative-competitive representation based classification}
\label{sec:sect_3}

\subsection{SCCRC method}
\label{sec:7}
Although CRC and its improved approaches achieve impressive results in various classification tasks, it does not mean that sparsity can be totally ignored. Deng \etal \cite{deng2013defense} pointed out that the dense coefficient of CRC would mislead the classification, and in a more recent work~\cite{deng2018face}, they find that when given uncontrolled and limited training data, the $\ell_{1}$-minimization technique obtains more desirable results than that of $\ell_{2}$-norm. Akhtar \etal \cite{akhtar2017efficient} also argued that sparsity plays an explicit role in accurate classification and they proposed a sparsity augmented collaborative representation based classification (SA-CRC) algorithm. Inspired by the above work, we present sparse and collaborative-competitive representation based classification (SCCRC) method which combines sparse and collaborative-competitive representation for classification. Concretely, we first obtain the coefficients of test sample by SRC and CCRC, respectively. Then the two coefficients are fused by element-wise multiplication. Finally, we classify the test sample to the class that has the minimal residual. The detailed procedures of our SCCRC is summarized in Algorithm \ref{alg1}.

\begin{algorithm} 
\caption{SCCRC} 
\label{alg1} 
\begin{algorithmic}[1]
\REQUIRE training data matrix $\mathbf{X}$, test data $\boldsymbol{y}$, parameter $\lambda$ for SRC, parameters $\lambda_{1}$ and $\lambda_{2}$ for CCRC.
\STATE Compute the sparse coefficient $\boldsymbol{\alpha}$ of test sample $\boldsymbol{y}$ according to Eq.~(\ref{equ1});
\STATE Obtain the coefficient of CCRC $\boldsymbol{\beta}$ for $\boldsymbol{y}$ based on Eq.~(\ref{equ10});
\STATE Compute the fused coefficient $\boldsymbol{f}=\boldsymbol{\alpha} \odot \boldsymbol{\beta}$;
\STATE Classify $\boldsymbol{y}$ to the class that has the least residual: $\textrm{identity}(\boldsymbol{y})=\underset{i}{\min}\left\|\boldsymbol{y}-\mathbf{X}_{i} \boldsymbol{f}_{i}\right\|_{2}$;
\ENSURE the identity of $\boldsymbol{y}$.
\end{algorithmic} 
\end{algorithm}

\subsection{{\color{red}Difference between SCCRC and CCRC-$\ell_1$}}
{\color{red}
Our proposed SCCRC aims to promote the sparsity of coefficient vector obtained by CCRC, which involves a simple multiplication of coefficient vectors derived by SRC and CCRC. Another intuitive way to increase the sparsity of coefficient vector of CCRC is modifying the objective function of CCRC, \ie, replacing the $\ell_2$-norm constraint on $\boldsymbol{\beta}$ with the $\ell_1$-norm. This method is called CCRC-$\ell_1$, and its objective function is formulated as follows,
\begin{equation}
\label{eq:obj_sccrcl1}
\underset{\boldsymbol{\beta}}{\textrm{min}}\left\{\|\boldsymbol{y}-\mathbf{X} \boldsymbol{\beta}\|_{2}^{2}+\lambda_{1}\|\boldsymbol{\beta}\|_{1}+\lambda_{2} \sum_{i=1}^{C}\left\|\boldsymbol{y}-\mathbf{X}_{i} \boldsymbol{\beta}_{i}\right\|_{2}^{2}\right\}
\end{equation}
To solve Eq.~(\ref{eq:obj_sccrcl1}), by introducing an auxiliary variable $\boldsymbol{z}$, Eq.~(\ref{eq:obj_sccrcl1}) can be converted into the following equivalent optimization problem,
\begin{equation}
\label{eq:equ_obj}
\underset{\boldsymbol{\beta},\boldsymbol{z}}{\textrm{min}}\left\{\|\boldsymbol{y}-\mathbf{X} \boldsymbol{\beta}\|_{2}^{2}+\lambda_{1}\|\boldsymbol{z}\|_{1}+\lambda_{2} \sum_{i=1}^{C}\left\|\boldsymbol{y}-\mathbf{X}_{i} \boldsymbol{\beta}_{i}\right\|_{2}^{2}\right\}, \ \textrm{s.t.} \ \boldsymbol{\beta}=\boldsymbol{z}
\end{equation}
The Augmented Lagrange Multiplier (ALM) scheme can be adopted to solve Eq.~(\ref{eq:equ_obj}), and the augmented Langrangian function is formulated as, 
\begin{equation}
\label{eq:lagrange}
\mathcal{L}(\boldsymbol{\beta},\boldsymbol{z},\boldsymbol{\theta},\mu)=\|\boldsymbol{y}-\mathbf{X} \boldsymbol{\beta}\|_{2}^{2}+\lambda_{1}\|\boldsymbol{z}\|_{1}+\lambda_{2} \sum_{i=1}^{C}\left\|\boldsymbol{y}-\mathbf{X}_{i} \boldsymbol{\beta}_{i}\right\|_{2}^{2}+\left \langle \boldsymbol{\theta},\boldsymbol{\beta}-\boldsymbol{z} \right \rangle+\frac{\mu}{2}\left \| \boldsymbol{\beta}-\boldsymbol{z}  \right \|_2^2
\end{equation}
where $\boldsymbol{\theta}$ is the Lagrange multiplier, and $\mu>0$ is a penalty parameter. Eq.~(\ref{eq:lagrange}) can be solved iteratively by updating $\boldsymbol{\beta}$ and $\boldsymbol{z}$ once at a time. The detailed procedures are presented as follows.

\textit{Update $\boldsymbol{\beta}$}: Fix $\boldsymbol{z}$ and update $\boldsymbol{\beta}$ by solving the following problem,
\begin{equation}
\label{eq:update_beta}
\underset{\boldsymbol{\beta}}{\textrm{min}} \ \|\boldsymbol{y}-\mathbf{X} \boldsymbol{\beta}\|_{2}^{2}+\lambda_{2} \sum_{i=1}^{C}\left\|\boldsymbol{y}-\mathbf{X}_{i} \boldsymbol{\beta}_{i}\right\|_{2}^{2}+\left \langle \boldsymbol{\theta},\boldsymbol{\beta}-\boldsymbol{z} \right \rangle+\frac{\mu}{2}\left \| \boldsymbol{\beta}-\boldsymbol{z}  \right \|_2^2
\end{equation}
Suppose $\mathbf{X}_i^{'}$ is a matrix that has the same size as $\mathbf{X}$, and $\mathbf{X}_i^{'}$ only consists of samples from the $i$-th class, \ie, $\mathbf{X}_i^{'}=[\boldsymbol{0},\ldots,\mathbf{X}_i,\ldots,\boldsymbol{0}]$, Equation~(\ref{eq:update_beta}) can be reformulated as,
\begin{equation}
\label{eq:equ_beta}
\underset{\boldsymbol{\beta}}{\textrm{min}} \ \|\boldsymbol{y}-\mathbf{X} \boldsymbol{\beta}\|_{2}^{2}+\lambda_{2} \sum_{i=1}^{C}\left\|\boldsymbol{y}-\mathbf{X}_{i}^{'} \boldsymbol{\beta}\right\|_{2}^{2}+\frac{\mu}{2}\left \| \boldsymbol{\beta}-\boldsymbol{z}+\frac{\boldsymbol{\theta}}{\mu}  \right \|_2^2
\end{equation}
which has the following closed-form solution,
\begin{equation}
\label{eq:solu_beta}
\boldsymbol{\beta}=\left [ \mathbf{X}^T\mathbf{X}+\lambda_2\sum_{i=1}^{C}(\mathbf{X}_i^{'})^{T}\mathbf{X}_i^{'}+\frac{\mu}{2}\mathbf{I} \right ]^{-1}((1+\lambda_2)\mathbf{X}^T\boldsymbol{y}+\frac{\mu \boldsymbol{z}-\boldsymbol{\theta}}{2})
\end{equation}
where $\mathbf{I}$ is an identity matrix.

\textit{Update $\boldsymbol{z}$}: Fix $\boldsymbol{\boldsymbol{\beta}}$ and update $\boldsymbol{z}$ by solving the following problem,
\begin{equation}
\label{eq:update_z}
\underset{\boldsymbol{z}}{\textrm{min}} \ \frac{\lambda_1}{\mu}\left \| \boldsymbol{z} \right \|_1+\frac{1}{2}\left \| \boldsymbol{z}-(\boldsymbol{\beta}+\frac{\boldsymbol{\theta}}{\mu}) \right \|
\end{equation}
Eq.~(\ref{eq:update_z}) can be solved by the soft-thresholding operator~\cite{combettes2005signal}. The complete procedures of solving Eq.~(\ref{eq:equ_obj}) are summarized in Algorithm~\ref{alg2}.}

\begin{algorithm} 
\caption{Solve Eq.~(\ref{eq:equ_obj}) via ALM}
\label{alg2} 
\begin{algorithmic}[1]
\REQUIRE Test sample $\boldsymbol{y}$, training data matrix $\mathbf{X}$, balancing parameter $\lambda_1$ and $\lambda_2$.
\STATE Initialize $\boldsymbol{z}=\boldsymbol{\theta}=\boldsymbol{0}$, $\mu=0.5$, $\mu_{\textrm{max}}=10^6$, $\rho=1.1$, $\varepsilon=10^{-3}$;
\WHILE{not converged} 
\STATE Update $\boldsymbol{\beta}$ by Eq.~(\ref{eq:solu_beta});
\STATE Update $\boldsymbol{z}$ by solving Eq.~(\ref{eq:update_z});
\STATE Update $\boldsymbol{\theta}$ by $\boldsymbol{\theta}=\boldsymbol{\theta}+\mu(\boldsymbol{\beta}-\boldsymbol{z})$;
\STATE Update $\mu$ by $\mu=\textrm{min}(\rho \mu,\mu_{\textrm{max}})$;
\STATE Check the convergence condiction: $\left \| \boldsymbol{\beta}-\boldsymbol{z} \right \|_{\infty}<\varepsilon$.
\ENDWHILE 
\ENSURE Coefficient vector $\boldsymbol{\beta}$.
\end{algorithmic} 
\end{algorithm}

{\color{red}From Eq.~(\ref{eq:obj_sccrcl1}), one can see that CCRC-$\ell_1$ directly imposes the $\ell_1$-norm constraint on the coefficient vector $\boldsymbol{\beta}$, and the $\ell_1$-norm constraint induces sparsity of $\boldsymbol{\beta}$. CCRC-$\ell_1$ is solved by ALM, which iteratively update the variables. By contrast, our proposed SCCRC is more straightforward, it fuses the coefficients of SRC and CCRC by multiplication. Note that coefficient vectors of test data can be viewed as some kind of features, and the multiplication fusion of coefficient vectors is equivalent to the data integration at the feature level. We believe that the feature level fusion can bring in improved performance, as demonstrated by the experimental results in Section~\ref{sec:sect_4}.}

\subsection{{\color{red}Rationale} of SCCRC}
\label{sec:8}
{\color{red}In our proposed SCCRC, the fused coefficient vector of a test sample $\boldsymbol{y}$ is derived by $\boldsymbol{f}=\boldsymbol{\alpha} \odot \boldsymbol{\beta}$, where $\boldsymbol{\alpha} $ and $\boldsymbol{\beta} $ are obtained by SRC and CCRC, respectively. According to the procedures of SRC and CCRC, the test sample $\boldsymbol{y}$ and training data matrix $\mathbf{X}$ are normalized to have unit $\ell_2$-norm. Therefore, absolute values of the entries in the coefficient vectors are less than 1. When both the elements in $\boldsymbol{\alpha}$ and $\boldsymbol{\beta}$ have large absolute value, after multiplication, the corresponding element in $\boldsymbol{f}$ will also have large absolute value. In most cases, coefficient vector obtained by SRC is sparse, while the coefficient vector obtained by CCRC is a little dense, and coefficients of the training samples, whose labels are the same as the test sample, tend to have large absolute value. Therefore, sparsity of the fused coefficient vector is promoted which is beneficial for the correct classification of the test sample.}

\begin{figure}[htbp]
  \centering
  \includegraphics[trim={0mm 0mm 0mm 0mm},clip, width = .6\textwidth]{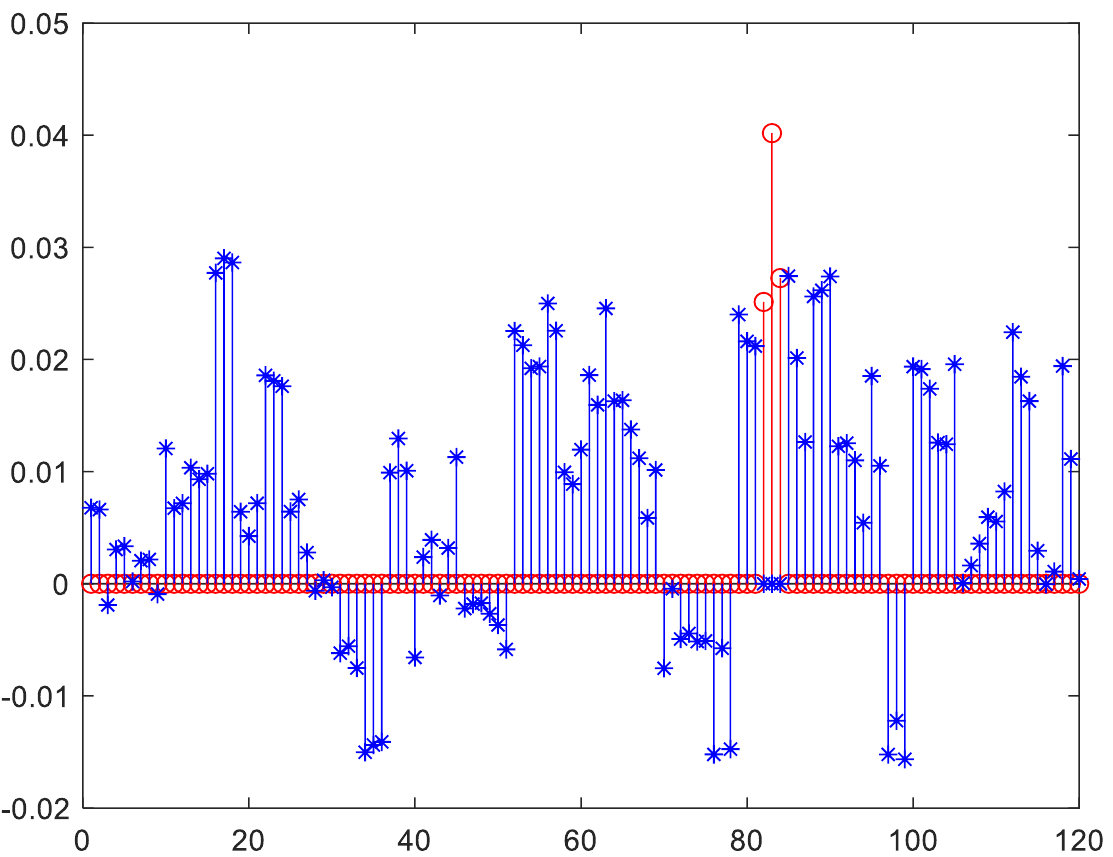}
  \caption{Coefficients computed by SRC, we can see that the prominent coefficients concentrate on the correct class (the 28th class).}
  \label{fig.1}
\end{figure}

\begin{figure}[htbp]
  \centering
  \includegraphics[trim={0mm 0mm 0mm 0mm},clip, width = .6\textwidth]{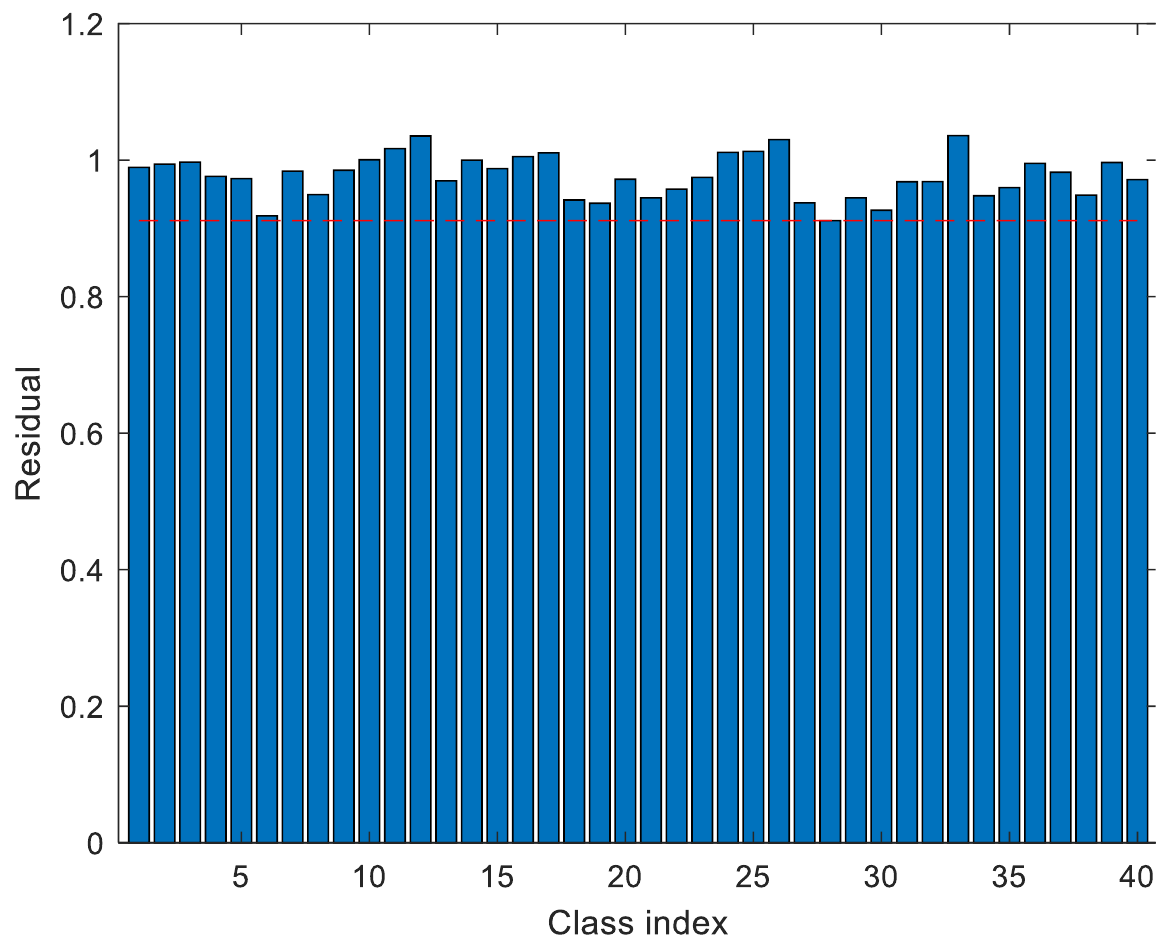}
  \caption{The residual obtained by SRC, and the 28th class has the least residual.}
  \label{fig.2}
\end{figure}

\begin{figure}[htbp]
  \centering
  \includegraphics[trim={0mm 0mm 0mm 0mm},clip, width = .6\textwidth]{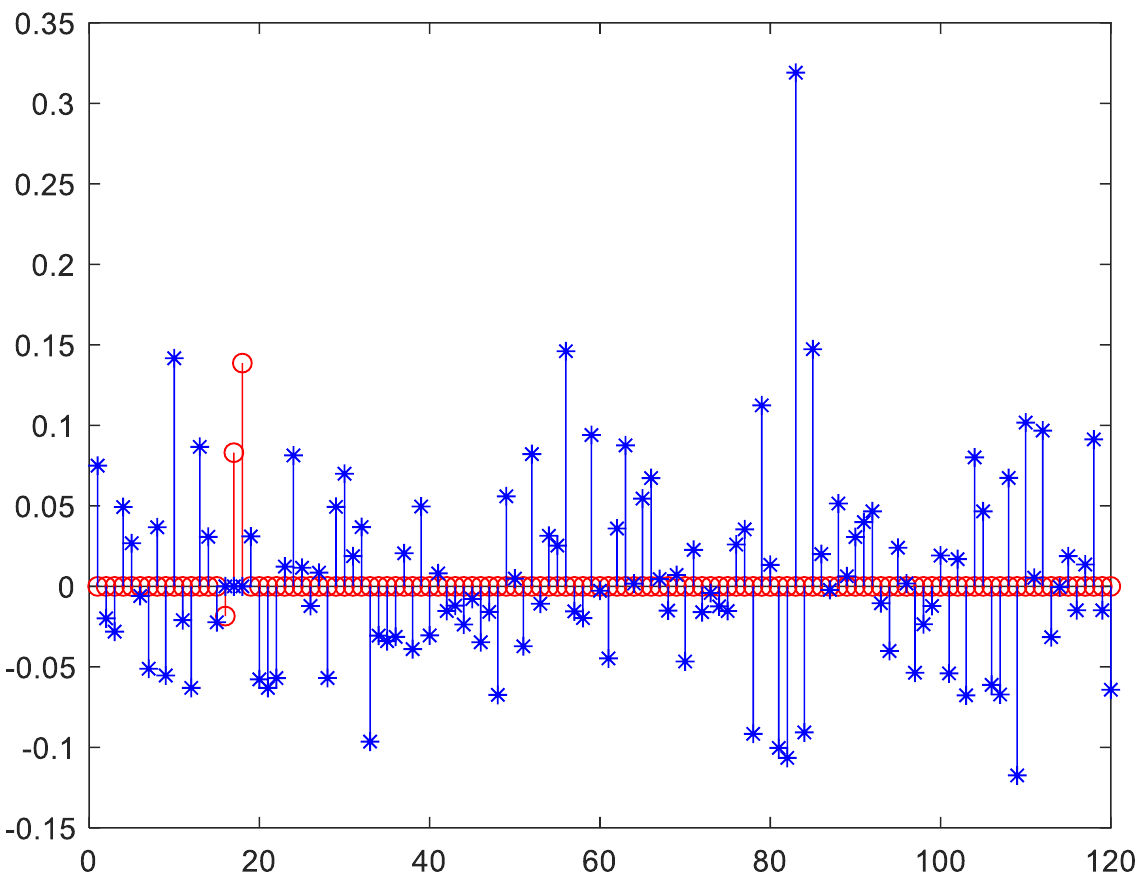}
  \caption{Coefficients computed by CCRC, although the 28th class has the largest coefficient, the other two coefficients are negative. In fact, the test sample is wrongly classified into the 6th class.}
  \label{fig.3}
\end{figure}

\begin{figure}[htbp]
  \centering
  \includegraphics[trim={0mm 0mm 0mm 0mm},clip, width = .6\textwidth]{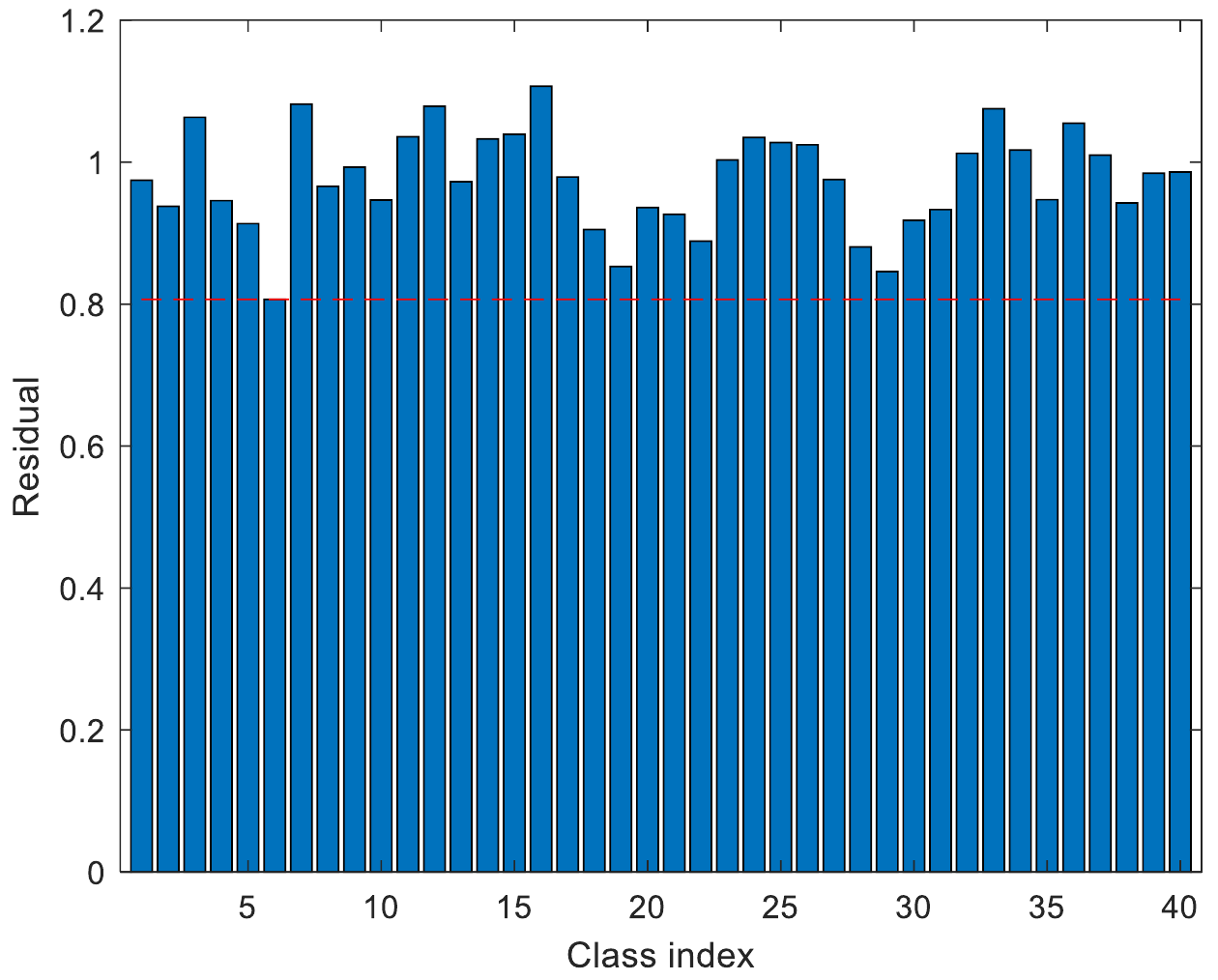}
  \caption{The residual obtained by CCRC, and the 6th class has the least residual. As a result, the test sample is misclassified.}
  \label{fig.4}
\end{figure}

\begin{figure}[htbp]
  \centering
  \includegraphics[trim={0mm 0mm 0mm 0mm},clip, width = .6\textwidth]{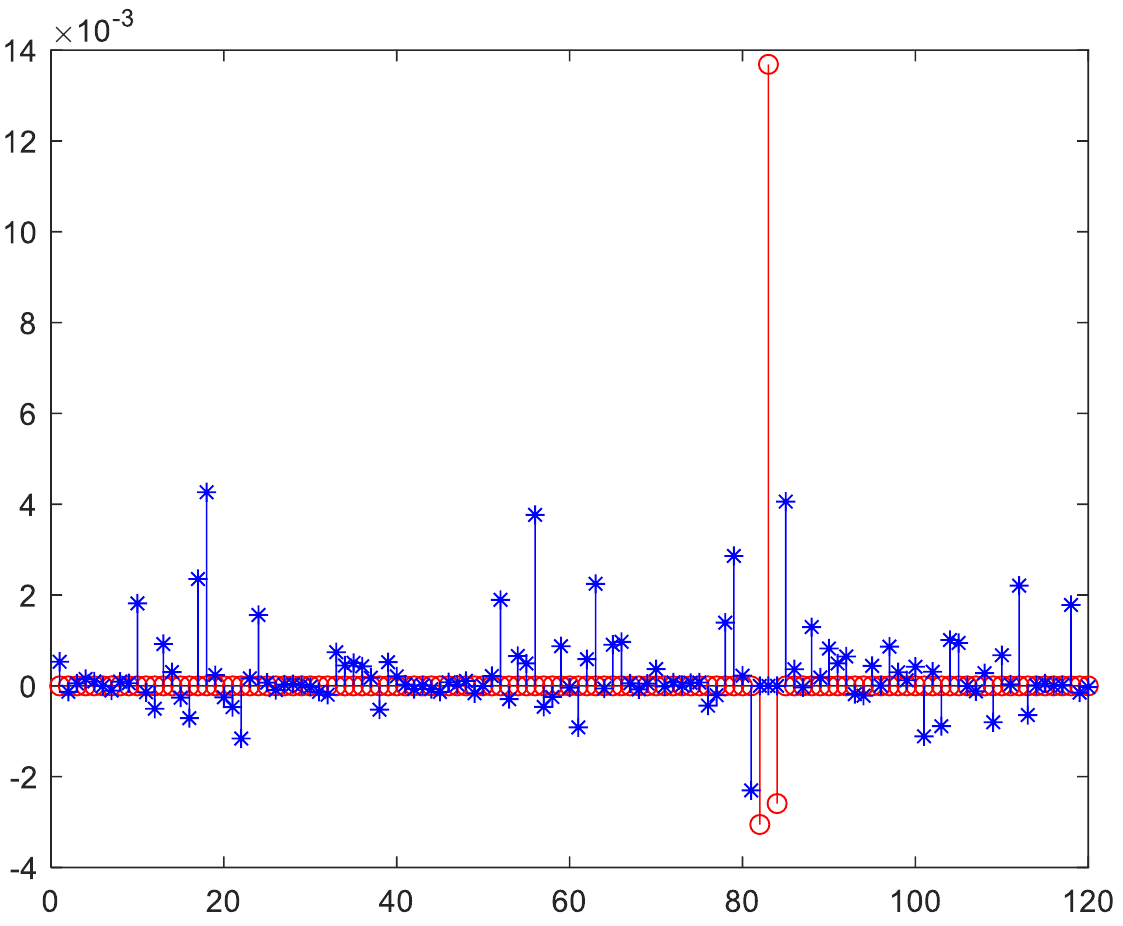}
  \caption{Coefficients computed by SCCRC, we can see that the dominant coefficients correspond to the correct class (the 28th class).}
  \label{fig.5}
\end{figure}

\begin{figure}[htbp]
  \centering
  \includegraphics[trim={0mm 0mm 0mm 0mm},clip, width = .6\textwidth]{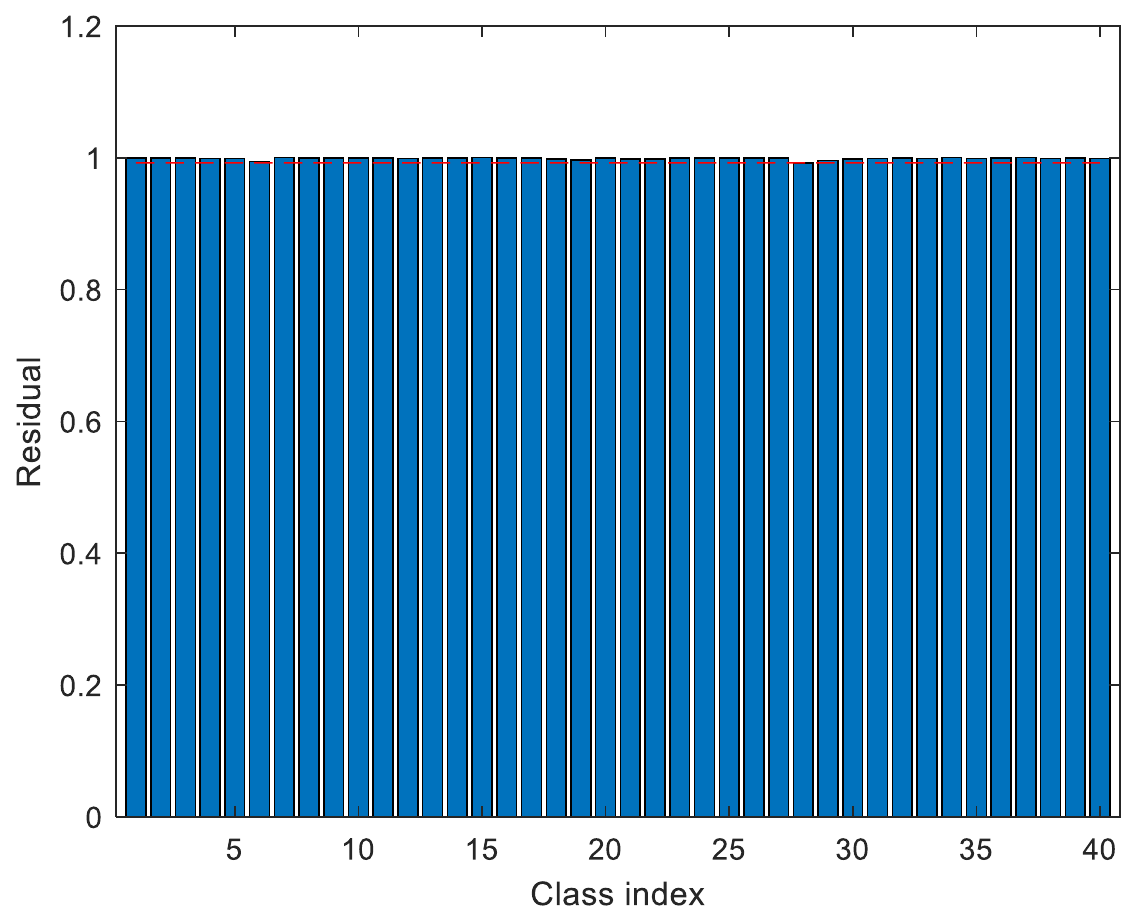}
  \caption{The residual obtained by SCCRC, and the 28th class has the least residual.}
  \label{fig.6}
\end{figure}

To vividly illustrate the effectiveness of SCCRC, we here present an example on the ORL database. We choose a test image from the 28th subject, and the training data consists of the first 3 images per person, and thus the dictionary contains $40\times3=120$ atoms. Fig. \ref{fig.1} shows the coefficients obtained by SRC, and the prominent coefficients correspond to the 28th class. Fig. \ref{fig.2} depicts the reconstruction error for each class, and the 28th class has the least residual, so the test sample is correctly classified. Fig. \ref{fig.3} plots the coefficients computed by CCRC, we can find that the largest coefficient corresponds to the 28th class. However, when we check the residual presented in Fig. \ref{fig.4}, the test sample is wrongly classified into the 6th class. The reason is that the other two coefficients of the 28th subject are negative, which leads to the fact that the 28th class does not produce the least residual. Fig. \ref{fig.5} shows the coefficients computed by our SCCRC, we can see that by fusing the coefficients of SRC and CCRC, the dominant coefficients correspond to the 28th class. Though the other two coefficients are negative, they are relatively small compared with the positive one. Thus, the residual of the 28th subject is the minimal, which is illustrated in Fig. \ref{fig.6}. {\color{red}By comparing Fig.~\ref{fig.5} with Figs.~\ref{fig.1} and~\ref{fig.3}, we can observe that the fused coefficient vector of SCCRC is more sparse than that of SRC and CCRC, which validates the effectiveness of multiplication fusion.}

{\color{red}From Figs.~\ref{fig.2},~\ref{fig.4} and~\ref{fig.6},} we can see that by augmenting the sparsity of coefficients obtained by CCRC, SCCRC can avoid the misclassification scenario to some degree. Next we employ another criteria to illustrate the effectiveness of SCCRC. In~\cite{wright2009robust}, Wright defined a measure of how concentrated the coefficients are on a single class, i.e. sparsity concentration index (SCI), which is defined as,

\begin{equation}\label{equ11}
\operatorname{SCI}(\boldsymbol{y})=\frac{C \cdot \underset{i}{\max} \left\|\boldsymbol{\alpha}_{i}\right\|_{1} /\|\boldsymbol{\alpha}\|_{1}-1}{C-1} \in[0,1]
\end{equation}

Bigger values of SCI means enhanced sparsity; therefore, we exploit this index to calculate the SCI of the above test sample. The values of SRC, CCRC and our proposed SCCRC are 0.0428, 0.0714 and 0.2079, respectively. We can see that by introducing the competitive regularization term, sparsity of CCRC is improved compared with SRC. By combining sparse and collaborative-competitive representation, our proposed SCCRC achieves the highest sparsity, thus improved classification results can be expected.

\section{Experimental results and analysis}
\label{sec:sect_4}
In this section, we report the performance of SCCRC on five publicly available datasets, i.e. ORL~\cite{samaria1994parameterisation}, Georgia Tech~\cite{goel2005face}, FERET~\cite{phillips1997feret}, Extended Yale B~\cite{georghiades2001few} and AR databases~\cite{martinez2007ar}. We compare the classification accuracy of SCCRC with SRC~\cite{wright2009robust}, LRC~\cite{naseem2010linear}, CRC~\cite{zhang2011sparse}, SCRC~\cite{zeng2017multiplication}, NRC~\cite{xu2019sparse}, ProCRC~\cite{cai2016probabilistic} CCRC~\cite{yuan2018collaborative} and CCRC-$\ell_1$. {\color{red}In addition, we present the classification time (in seconds) of all the competing approaches.} We use SolveFISTA.m~\cite{yang2010fast} to solve the sparse optimization problem. The parameters $\lambda_1$ and $\lambda_2$ in CCRC and our proposed SCCRC are selected from the set $\left\{10^{-7}, 10^{-6}, \cdots, 1,10,100\right\}$. All experiments are conducted with MATLAB R2019b under Windows 10 on a PC equipped with Intel i9-8950HK 2.90 GHz CPU and 32 GB RAM.

\subsection{Experiments on the ORL database}
\label{sec:4-1}
\begin{figure}[htbp]
  \centering
  \includegraphics[trim={0mm 0mm 0mm 0mm},clip, width = .6\textwidth]{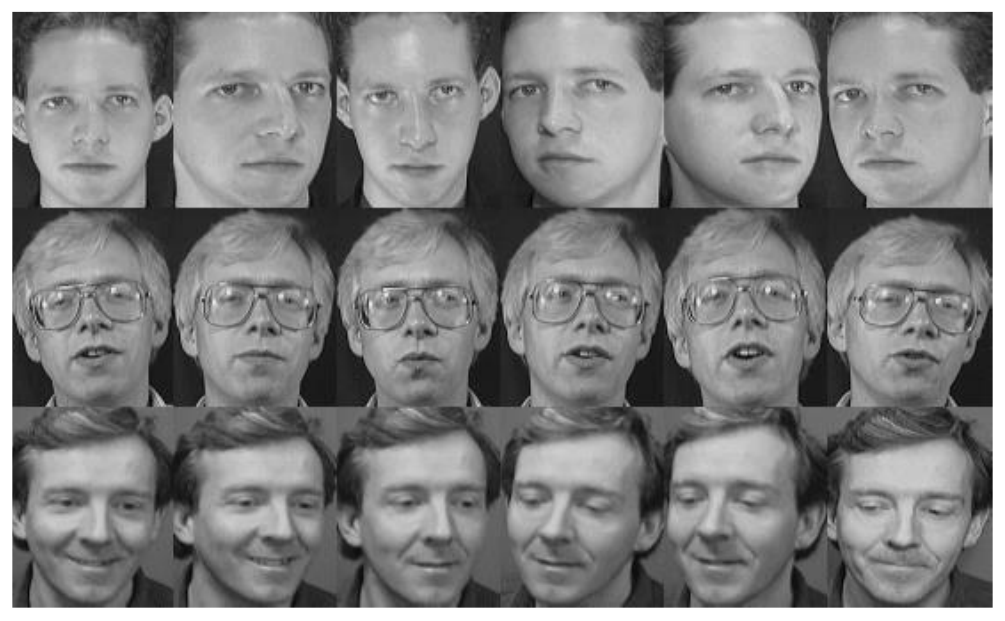}
  \caption{Example images from the ORL database.}
  \label{fig.7}
\end{figure}

The ORL database contains 400 images of 40 individuals. For each subject, there are 10 images with the variations in lighting, facial expression and facial details (with or without glasses), Fig. \ref{fig.7} shows some face images from this database. In our experiments, each image is resized to 56$\times$46, and the first 1 to 6 images per person are selected as training samples and the remaining are testing samples.

{\color{red}The recognition accuracy and the testing time (when the first 6 images per subject are used as training samples) of different methods are shown in Table \ref{tab:resu_orl}.} We can see that SCCRC outperforms all the competing approaches {\color{red}in terms of recognition accuracy. Meanwhile, the testing time of SCCRC is comparable to that of SCRC, and it is 31 times faster than NRC}. With the increase of the number of training samples, the classification accuracy of SCCRC increases consistently. By integrating SRC and CRC, SCRC obtains better results than those of SRC and CRC, especially when the number of training samples is relatively small. By introducing the non-negative constraint, NRC exhibits superiority over SRC and CRC. Thanks to the competitive regularization term, CCRC outperforms CRC with the increasing number of training samples.

\begin{table}[]
\caption{Classification accuracy (\%) and the testing time of competing approaches on the ORL database.}
\label{tab:resu_orl}
\centering
\begin{tabular}{cccccccc}
\hline
Methods & 1              & 2              & 3              & 4              & 5              & 6     & {\color{red}testing time (s)}         \\ \hline
SRC{\color{red}~\cite{wright2009robust}}     & 73.06          & 82.50          & 84.64          & 88.33          & 88.50          & 92.50   & {\color{red}0.79 }      \\
LRC{\color{red}~\cite{naseem2010linear}}     & 67.50          & 79.38          & 81.43          & 86.25          & 88.00          & 94.38   & {\color{red}0.70 }       \\
CRC{\color{red}~\cite{zhang2011sparse}}     & 71.67          & 83.75          & 86.07          & 91.25          & 90.50          & 92.50    & {\color{red}0.34 }      \\
SCRC{\color{red}~\cite{zeng2017multiplication}}    & \textbf{73.89} & 86.56          & 86.43          & 91.25          & 92.50          & 92.50  & {\color{red}0.77 }        \\
NRC{\color{red}~\cite{xu2019sparse}}     & 72.78          & 87.19          & 88.21          & 90.00          & 90.00          & 93.75  & {\color{red}28.66 }        \\
ProCRC{\color{red}~\cite{cai2016probabilistic}}  & 67.50          & 85.00          & 86.07          & 90.41          & 92.00          & 94.37   & {\color{red}0.09 }       \\
CCRC{\color{red}~\cite{yuan2018collaborative}}    & 68.06          & 85.63          & 87.50          & 90.83          & 92.50          & 94.38     & {\color{red}0.29 }     \\
{\color{red}CCRC-$\ell_1$}    &  {\color{red}71.67}          & {\color{red}85.63}           & {\color{red}87.14}          & {\color{red}90.00}          & {\color{red}89.50}          & {\color{red}93.13}      & {\color{red}0.29 }     \\
SCCRC   & \textbf{73.89} & \textbf{87.81} & \textbf{88.57} & \textbf{91.67} & \textbf{93.00} & \textbf{95.00} & {\color{red}0.91 }\\ \hline
\end{tabular}
\end{table}

\subsection{Experiments on the GT database}
\label{sec:4-2}
\begin{figure}[htbp]
  \centering
  \includegraphics[trim={0mm 0mm 0mm 0mm},clip, width = .6\textwidth]{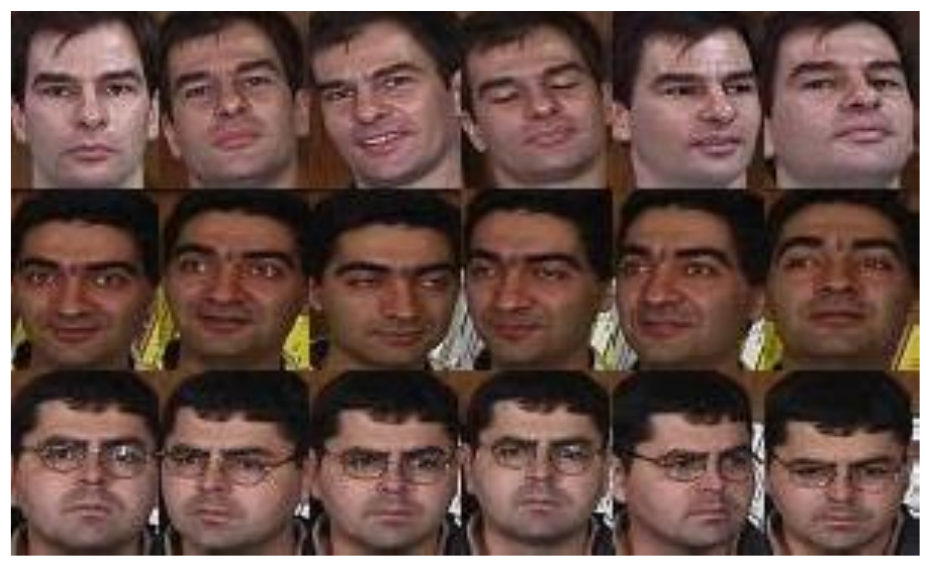}
  \caption{Example images from the GT database.}
  \label{fig.8}
\end{figure}

There are 750 face images (50 individuals and each has 15 images) in the GT face database. These face images show frontal and/or tilted faces with different facial expressions, lighting conditions and scale. In our experiments, all images are cropped and resized to 60$\times$50. Fig. \ref{fig.8} shows some cropped face images from the GT face dataset. The first 1-6 face images of each subject are used as training samples, and the remaining images are regarded as testing samples.

{\color{red}Table \ref{tab:resu_gt} lists the classification accuracy and the testing time (when the first 6 images per subject are used as training samples) of compared methods on the GT database.} One can see that with the increasing number of training samples, accuracy of all competing methods is increased, and the proposed SCCRC achieves the best results in most cases. {\color{red}Moreover, ProCRC has the shortest testing time, and SCCRC is 25 times faster than NRC.}

\begin{table}[]
\caption{Classification accuracy (\%) and the testing time of various approaches on the GT database.}
\label{tab:resu_gt}
\centering
\begin{tabular}{cccccccc}
\hline
Methods & 1              & 2              & 3              & 4              & 5              & 6      & {\color{red}testing time (s)}          \\ \hline
SRC{\color{red}~\cite{wright2009robust}}      & 38.71          & 47.38          & 49.17          & 52.18          & 53.60          & 62.44   & {\color{red}5.38 }        \\
LRC{\color{red}~\cite{naseem2010linear}}      & 36.14          & 46.62          & 51.33          & 55.64          & 59.60          & 67.33    & {\color{red}2.81 }       \\
CRC{\color{red}~\cite{zhang2011sparse}}      & 36.43          & 46.00          & 49.33          & 54.36          & 58.40          & 64.67    & {\color{red}1.68 }       \\
SCRC{\color{red}~\cite{zeng2017multiplication}}     & 37.71          & 48.31          & 53.83          & 56.91          & 61.20          & 68.00     & {\color{red}4.47 }      \\
NRC{\color{red}~\cite{xu2019sparse}}      & 36.86          & 47.69          & 50.33          & \textbf{57.82} & 60.20          & 66.89 & {\color{red}116.20 }          \\
ProCRC{\color{red}~\cite{cai2016probabilistic}}   & 34.14          & 45.69          & 51.33          & 55.09          & 58.40          & 63.77     & {\color{red}0.35 }      \\
CCRC{\color{red}~\cite{yuan2018collaborative}}     & 33.43          & 44.46          & 48.67          & 51.64          & 55.40          & 61.33      & {\color{red}1.30 }     \\
{\color{red}CCRC-$\ell_1$}    &  {\color{red}33.86}          & {\color{red}44.77}           & {\color{red}48.67}          & {\color{red}51.64}          & {\color{red}55.20}          & {\color{red}62.89}      & {\color{red}1.32 }     \\
SCCRC   & \textbf{39.14} & \textbf{50.15} & \textbf{54.50} & 57.64          & \textbf{62.20} & \textbf{68.89}& {\color{red}4.57 }  \\ \hline
\end{tabular}
\end{table}

\subsection{Experiments on the FERET database}
\label{sec:4-3}
\begin{figure}[htbp]
  \centering
  \includegraphics[trim={0mm 0mm 0mm 0mm},clip, width = .6\textwidth]{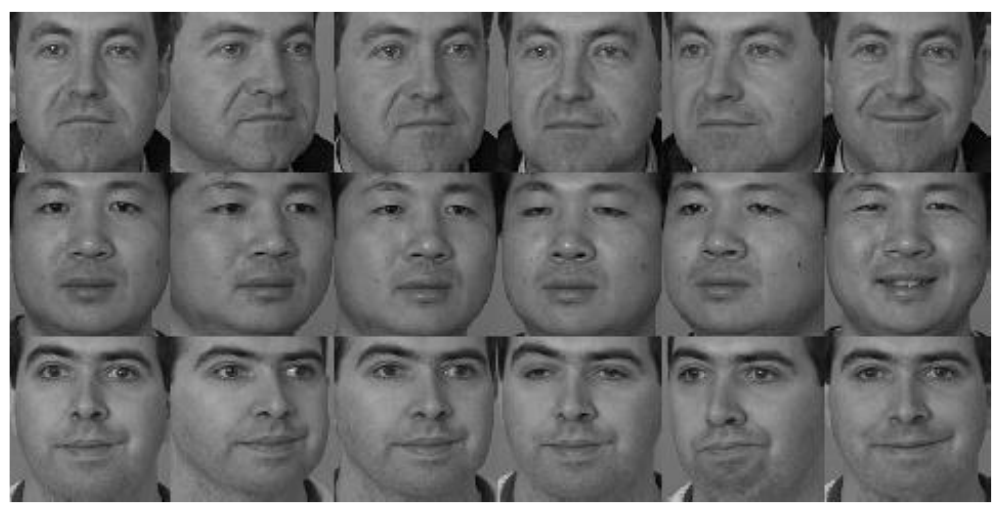}
  \caption{Example images from the FERET database.}
  \label{fig.9}
\end{figure}

This subset used is from the well-known FERET face database, which consists of 1400 face images of 200 subjects, each provided with seven different face images. This subset is composed of images in the original FERET face data set whose names are marked with two-character strings: ba, bj, bk, be, bf, bd, and bg. Fig. \ref{fig.9} shows some of these face images. Each face image is resized to a 40$\times$40 image. The first 1-6 face images of each subject are selected as training samples, and the rest as testing samples.

Table \ref{tab:resu_feret} details the classification results of competing methods. {\color{red}One can see that SCCRC is very efficient, and its testing time is only one-fifth of that of NRC.} The classification results of all approaches drop when the number of training samples per subject is 3, ~\cite{zeng2017multiplication} also reported this observation. The reason behind this phenomenon is that we select the first 3 images for all subjects, that is, the indices of training samples are fixed across different individuals. A more reasonable way is to randomly choose the training samples for each subject.

\begin{table}[]
\caption{Classification accuracy (\%) and the testing time of different methods on the FERET database.}
\label{tab:resu_feret}
\centering
\begin{tabular}{cccccccc}
\hline
Methods & 1              & 2              & 3              & 4              & 5              & 6     & {\color{red}testing time (s)}         \\ \hline
SRC{\color{red}~\cite{wright2009robust}}     & 45.83          & 57.20          & 52.38          & 68.00          & 76.25          & 79.50  & {\color{red}8.57 }         \\
LRC{\color{red}~\cite{naseem2010linear}}     & 44.50          & 63.10          & 58.88          & \textbf{76.33} & 81.50          & 75.50    & {\color{red}3.36 }       \\
CRC{\color{red}~\cite{zhang2011sparse}}     & 40.67          & 55.60          & 47.38          & 56.83          & 65.25          & 69.50   & {\color{red}12.96 }        \\
SCRC{\color{red}~\cite{zeng2017multiplication}}    & 45.67          & 61.90          & 56.00          & 69.17          & 80.50          & 80.50     & {\color{red}8.61 }      \\
NRC{\color{red}~\cite{xu2019sparse}}     & 46.33          & 59.90          & 52.62          & 63.00          & 73.75          & 73.00  & {\color{red}42.46 }         \\
ProCRC{\color{red}~\cite{cai2016probabilistic}}  & 39.75          & 54.90          & 42.87          & 49.66          & 56.75          & 50.00     & {\color{red}0.43 }      \\
CCRC{\color{red}~\cite{yuan2018collaborative}}    & 42.08          & 57.20          & 48.25          & 53.83          & 57.00          & 47.00     & {\color{red}5.59 }      \\
{\color{red}CCRC-$\ell_1$}    &  {\color{red}43.42}          & {\color{red}58.50}           & {\color{red}48.63}          & {\color{red}59.33}          & {\color{red}69.75}          & {\color{red}68.50}      & {\color{red}6.09 }     \\
SCCRC   & \textbf{47.83} & \textbf{65.50} & \textbf{60.12} & 74.83          & \textbf{83.25} & \textbf{85.50}& {\color{red}8.92 }  \\ \hline
\end{tabular}
\end{table}

{\color{red}In order to demonstrate the statistical significance of our proposed SCCRC compared with the other methods, we conduct a significance test, McNemar's test~\cite{wen2018low,li2015learning}, for the results shown in Table~\ref{tab:resu_feret}. The significance level, i.e., $p$-value is set as 0.05, which means that the performance difference between two methods is statistically significant, if the estimated $p$-value is lower than 0.05. Table~\ref{tab:sig_test} lists the $p$-values between SCCRC and the other methods. From this table, one can see that the performance differences between SCCRC and the methods (SRC, CRC, SCRC, ProCRC, CCRC and CCRC-$\ell_1$) are statistically significant in all cases. The performance differences between SCCRC and LRC/NRC are not statistically significant; however, SCCRC outperforms NRC in all cases, and SCCRC is superior to LRC except when there are 4 training samples per subject. The above experimental results validate the effectiveness of our proposed SCCRC.}

\begin{table}[]
\caption{{\color{red}$p$-value between SCCRC and the other methods on the FERET database. $*$ indicates that the difference between the two methods is statistically significant when $p$=0.05.}}
\label{tab:sig_test}
\centering
\begin{tabular}{ccccccc}
\hline
{\color{red}Methods} & {\color{red}1}  & {\color{red}2}  & {\color{red}3}              & {\color{red}4} & {\color{red}5}& {\color{red}6} \\ \hline
{\color{red}SRC~\cite{wright2009robust}}     &     {\color{red}$0.0165^{*}$}    &    {\color{red}6.59$\times 10^{-12*}$}      & {\color{red}2.54$\times 10^{-8*}$} &  {\color{red}2.73$\times 10^{-6*}$}  &  {\color{red}2.64$\times 10^{-4*}$}   &   {\color{red}$0.0146^{*}$}      \\
{\color{red}LRC~\cite{naseem2010linear}}     &   {\color{red}$0.0011^{*}$}       & {\color{red}$0.0399^{*}$}       & {\color{red}0.3291} & {\color{red}0.2370} &    {\color{red}0.3222}     &  {\color{red}1.03$\times 10^{-4*}$}    \\
{\color{red}CRC~\cite{zhang2011sparse}}     &    {\color{red}3.90$\times 10^{-9*}$}        &  {\color{red}1.05$\times 10^{-14*}$} & {\color{red}1.83$\times 10^{-16*}$} &    {\color{red}5.67$\times 10^{-18*}$}      &    {\color{red}2.52$\times 10^{-16*}$}  &  {\color{red}1.02$\times 10^{-7*}$}  \\
{\color{red}SCRC~\cite{zeng2017multiplication}}    &   {\color{red}$0.0271^{*}$}   & {\color{red}2.18$\times 10^{-4*}$} & {\color{red}3.74$\times 10^{-4*}$} & {\color{red}7.27$\times 10^{-4*}$} &   {\color{red}$0.0227^{*}$}       & {\color{red}$0.0127^{*}$}         \\
{\color{red}NRC~\cite{xu2019sparse}}     &   {\color{red}0.1783}        &  {\color{red}6.71$\times 10^{-7*}$}        & {\color{red}5.68$\times 10^{-8*}$} & {\color{red}1.13$\times 10^{-8*}$} &   {\color{red}4.00$\times 10^{-8*}$}       & {\color{red}8.67$\times 10^{-7*}$}       \\
{\color{red}ProCRC~\cite{cai2016probabilistic}}  &    {\color{red}1.79$\times 10^{-10*}$}     & {\color{red}6.32$\times 10^{-16*}$}     & {\color{red}1.28$\times 10^{-23*}$} & {\color{red}1.46$\times 10^{-26*}$} & {\color{red}2.58$\times 10^{-23*}$} &   {\color{red}5.23$\times 10^{-19*}$     } \\
{\color{red}CCRC~\cite{yuan2018collaborative}}    &  {\color{red}5.55$\times 10^{-6*}$}        & {\color{red}2.93$\times 10^{-10*}$}        & {\color{red}2.74$\times 10^{-13*}$}  &  {\color{red}4.65$\times 10^{-19*}$}        &    {\color{red}2.20$\times 10^{-21*}$}      &   {\color{red}1.02$\times 10^{-20*}$}    \\ 
{\color{red}CCRC-$\ell_1$}    &  {\color{red}1.77$\times 10^{-4*}$}        & {\color{red}1.32$\times 10^{-8*}$}        & {\color{red}5.26$\times 10^{-14*}$}  &  {\color{red}1.70$\times 10^{-12*}$}        &    {\color{red}1.49$\times 10^{-12*}$}      &   {\color{red}3.62$\times 10^{-7*}$}    \\\hline
\end{tabular}
\end{table}

\subsection{Experiments on the Extended Yale B database}
\label{sec:4-4}
\begin{figure}[htbp]
  \centering
  \includegraphics[trim={0mm 0mm 0mm 0mm},clip, width = .6\textwidth]{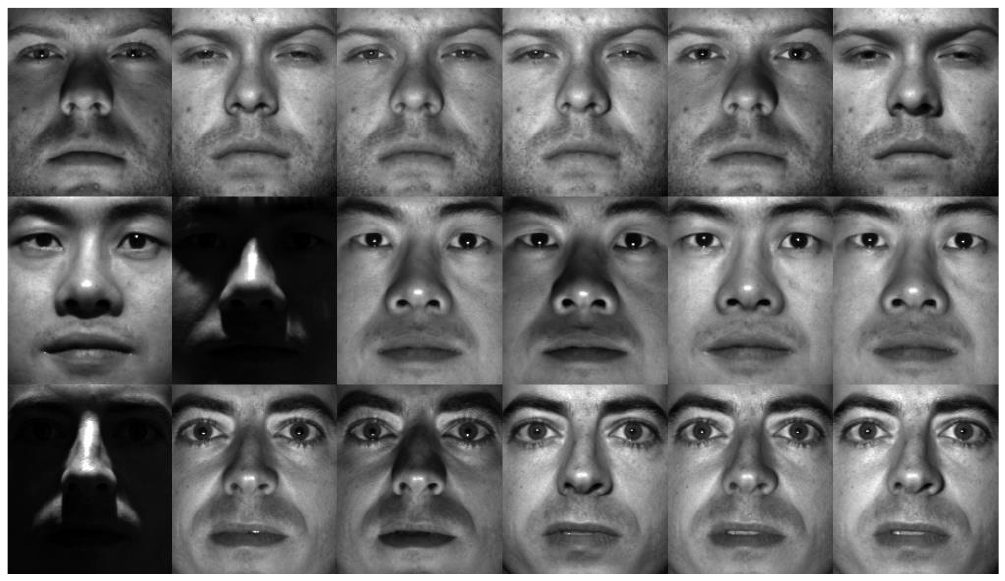}
  \caption{Example images from the Extended Yale B database.}
  \label{fig.10}
\end{figure}

The Extended Yale B~\cite{georghiades2001few} face dataset contains 2414 frontal facial images of 38 individuals. These images are captured under various controlled lighting conditions. The size of an image is 192$\times$168 pixels. In our experiments, all images are cropped and resized to 32$\times$32 pixels. Fig. \ref{fig.10} shows some face images from the Extended Yale B face dataset. The first 5, 10, 15, 20, 25 and 30 face images of each subject are treated as training samples and the remaining as testing samples.

{\color{red}Classification accuracy and the testing time (when the first 30 images per subject are used as training samples) of different approaches on this database is listed in Table \ref{tab:resu_yaleb}. We can observe that the testing time of SCCRC is comparable to that of SCRC, and it is about 3 times faster than NRC.} Similar to the experimental results on the above three databases, classification accuracy of SCCRC increases steadily with the increasing number of training samples per subject, and the performance gain is significant when the number of training samples is increasing. For example, when the number of training samples is 15, SCCRC has more than 6.73\% higher accuracy than the second best method, \ie, LRC.

\begin{table}[]
\caption{Classification accuracy (\%) and the testing time of competing algorithms on the Extended Yale B database.}
\label{tab:resu_yaleb}
\centering
\begin{tabular}{cccccccc}
\hline
Methods & 5              & 10             & 15             & 20             & 25             & 30       & {\color{red}testing time (s)}     \\ \hline
SRC{\color{red}~\cite{wright2009robust}}     & 40.51          & 41.79          & 38.77          & 41.17          & 44.81          & 44.82  & {\color{red}38.82 }       \\
LRC{\color{red}~\cite{naseem2010linear}}     & 46.27          & 75.22          & 75.27          & 76.06          & 83.06          & 85.01  & {\color{red}14.69 }        \\
CRC{\color{red}~\cite{zhang2011sparse}}     & 48.29          & 68.24          & 74.13          & 77.81          & 79.85          & 85.09  & {\color{red}5.45 }       \\
SCRC{\color{red}~\cite{zeng2017multiplication}}    & 46.09          & 57.03          & 65.94          & 69.11          & 67.62          & 71.82     & {\color{red}41.59 }     \\
NRC{\color{red}~\cite{xu2019sparse}}     & 51.08          & 65.93          & 71.04          & 74.12          & 77.12          & 81.63 & {\color{red}113.94 }       \\
ProCRC{\color{red}~\cite{cai2016probabilistic}}  & 52.02          & 69.27          & 73.69          & 77.14          & 80.94          & 85.63   & {\color{red}0.36 }      \\
CCRC{\color{red}~\cite{yuan2018collaborative}}    & \textbf{54.99} & 72.62          & 74.95          & 77.33          & 83.06          & 85.56   & {\color{red}2.54 }      \\
{\color{red}CCRC-$\ell_1$}    &  {\color{red}51.89}          & {\color{red}71.14}           & {\color{red}75.05}          & {\color{red}78.66}          & {\color{red}85.25}          & {\color{red}89.80}      & {\color{red}5.86 }     \\
SCCRC   & 52.97          & \textbf{79.25} & \textbf{82.00} & \textbf{83.56} & \textbf{83.81} & \textbf{87.21}& {\color{red}40.41 } \\ \hline
\end{tabular}
\end{table}

\subsection{Experiments on the AR database}
\label{sec:4-5}
\begin{figure}[htbp]
  \centering
  \includegraphics[trim={0mm 0mm 0mm 0mm},clip, width = .6\textwidth]{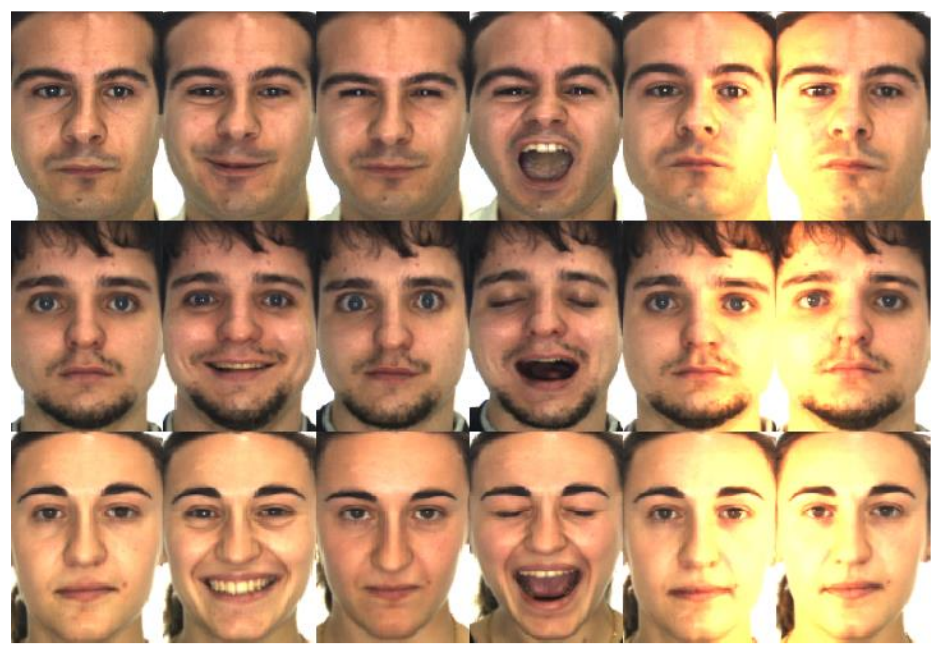}
  \caption{Example images from the AR database.}
  \label{fig.11}
\end{figure}

AR database includes over 4000 face images of 126 people (70 male and 56 female) which vary in expression, illumination and disguise (wearing sunglasses or scarves). Each subject has 26 images consisting of 14 clean images, 6 images with sunglasses and 6 images with scarves. As in~\cite{jiang2013label,yuan2018collaborative}, we use a subset that contains 1400 clean faces selected from 50 male and 50 female subjects, all the images are resized to 28$\times$20, and some example images are shown in Fig. \ref{fig.11}. For each subject, we use the first 1-6 face images as training samples, and the remaining as testing samples. {\color{red}Experimental results are shown in Table \ref{tab:resu_ar}.} Once again, SCCRC outperforms the other competing approaches {\color{red}in terms of classification accuracy} except for the case of 2 training samples per person.

\begin{table}[]
\caption{Classification accuracy (\%) and the testing time of evaluating methods on the AR database.}
\label{tab:resu_ar}
\centering
\begin{tabular}{cccccccc}
\hline
Methods & 1              & 2              & 3              & 4              & 5              & 6      & {\color{red}testing time (s)}         \\ \hline
SRC{\color{red}~\cite{wright2009robust}}      & 59.00          & 56.33          & 53.91          & 52.90          & 55.67          & 61.12  & {\color{red}4.19 }        \\
LRC{\color{red}~\cite{naseem2010linear}}      & 58.08          & 56.92          & 55.64          & 56.50          & 60.00          & 75.50     & {\color{red}5.06 }     \\
CRC{\color{red}~\cite{zhang2011sparse}}      & 58.15          & 60.42          & 61.09          & 65.70          & 76.89          & 91.50    & {\color{red}2.91 }      \\
SCRC{\color{red}~\cite{zeng2017multiplication}}     & 62.08          & 64.17          & 65.00          & 66.70          & 75.78          & 88.62   & {\color{red}4.48 }       \\
NRC{\color{red}~\cite{xu2019sparse}}      & 65.31          & 68.33          & 68.64          & 72.60          & 84.00          & 93.00   & {\color{red}17.41 }       \\
ProCRC{\color{red}~\cite{cai2016probabilistic}}   & 68.46          & \textbf{73.00} & 75.72          & 75.70          & 84.00          & 91.62   & {\color{red}0.28 }       \\
CCRC{\color{red}~\cite{yuan2018collaborative}}     & 67.62          & 71.92          & 72.82          & 73.20          & 83.00          & 91.25   & {\color{red}1.17 }       \\
{\color{red}CCRC-$\ell_1$}    &  {\color{red}63.77}          & {\color{red}68.75}           & {\color{red}70.36}          & {\color{red}73.30}          & {\color{red}85.11}          & {\color{red}93.13}      & {\color{red}1.49 }     \\
SCCRC   & \textbf{70.00} & 72.75          & \textbf{77.27} & \textbf{78.90} & \textbf{86.11} & \textbf{93.63}& {\color{red}4.54 } \\ \hline
\end{tabular}
\end{table}

\subsection{{\color{red}Experiments on the corrupted face images}}
\label{sec:4-6}
\begin{figure}[htbp]
  \centering
  \includegraphics[trim={0mm 0mm 0mm 0mm},clip, width = .6\textwidth]{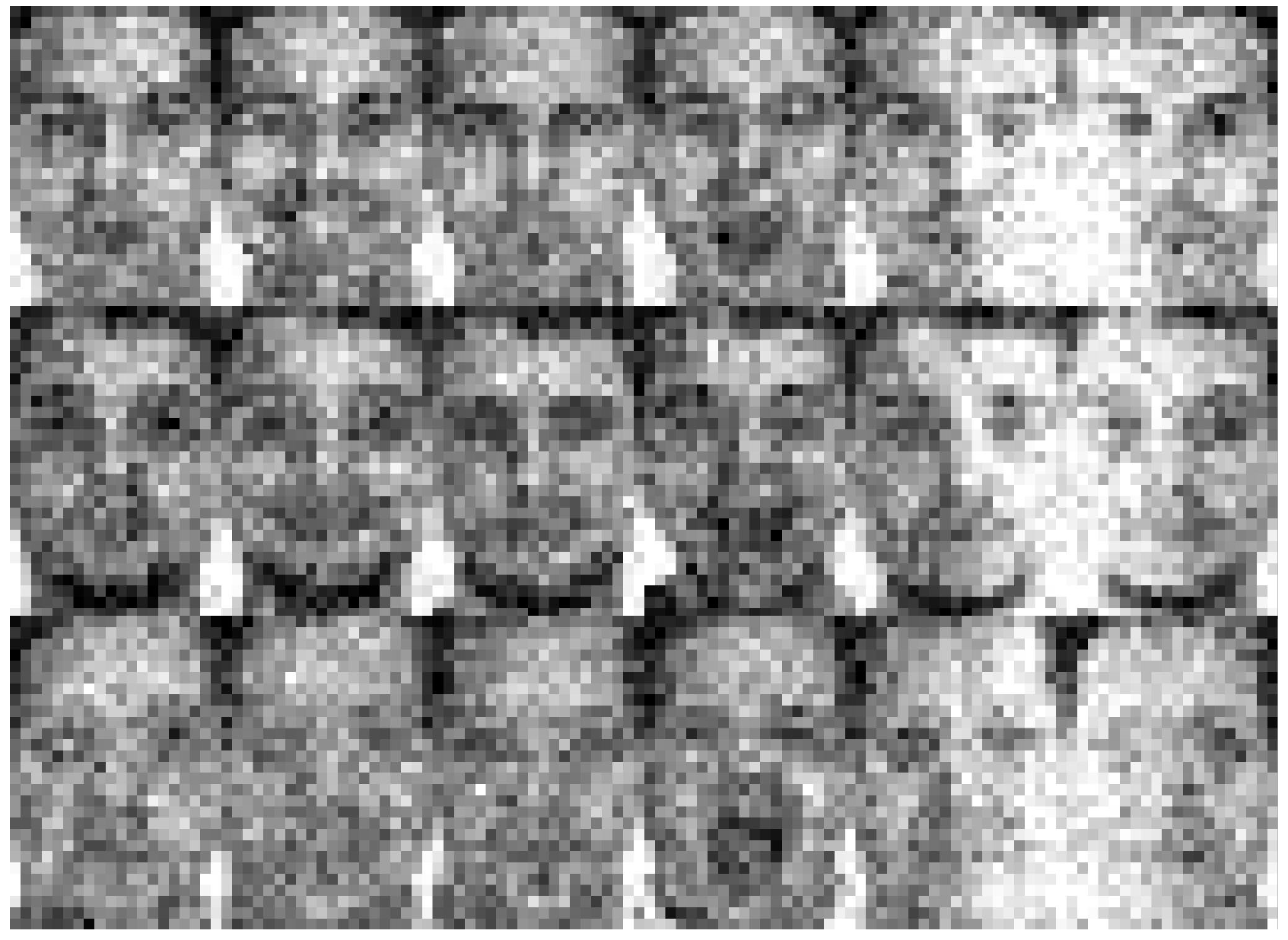}
  \caption{{\color{red}Some corrupted test images from the AR database.}}
  \label{fig:corr_imgs}
\end{figure}

{\color{red}To explore the robustness of our proposed method to noise, we use corrupted face images as test data. Here the AR database is used for evaluation, as in Section~\ref{sec:4-5}, 1400 images of 100 subjects are selected, and the size of image is 28$\times$20. The first seven images are used as training samples and the remaining as test samples. We add zero-mean, Gaussian white noise with variance of 0.01 to all the test images, some corrupted test face images are shown in Fig.~\ref{fig:corr_imgs}. Experimental results of all competing approaches are shown in Table~\ref{tab:corr_ar}. It can be seen that by introducing the $\ell_1$-norm constraint on the coefficient vector, CCRC-$\ell_1$ outperforms CCRC by 10.14\%. Our proposed SCCRC achieves the highest accuracy, and it is around 2.4 times faster than NRC.}

\begin{table}[]
\caption{{\color{red}Classification accuracy and the testing time of evaluating methods on the corrupted face images.}}
\label{tab:corr_ar}
\centering
\begin{tabular}{ccc}
\hline
{\color{red}Methods} & {\color{red}Accuracy (\%)}   & {\color{red}testing time (s)}                    \\ \hline
{\color{red}SRC~\cite{wright2009robust}}      &   {\color{red}84.57}   &   {\color{red}10.82}     \\
{\color{red}LRC~\cite{naseem2010linear}}      &   {\color{red}74.57}    &   {\color{red}4.75}   \\
{\color{red}CRC~\cite{zhang2011sparse}}      &      {\color{red}82.14}   &   {\color{red}2.80}    \\
{\color{red}SCRC~\cite{zeng2017multiplication}}    &  {\color{red}84.00}    &  {\color{red}4.84}    \\
{\color{red}NRC~\cite{xu2019sparse}}      &    {\color{red}86.14}        &   {\color{red}26.85}  \\
{\color{red}ProCRC~\cite{cai2016probabilistic}}   &   {\color{red}83.14}     & {\color{red}0.29}   \\
{\color{red}CCRC~\cite{yuan2018collaborative}}     &  {\color{red}75.00}      &  {\color{red}1.25} \\
{\color{red}CCRC-$\ell_1$}    &  {\color{red}85.14}          & {\color{red}1.65}         \\
{\color{red}SCCRC}   &   {\color{red}\textbf{87.29}}   &   {\color{red}11.27}  \\ \hline
\end{tabular}
\end{table}

\subsection{{\color{red}Experiment analysis}}
\label{sec:4-7}
{\color{red}
The classification results on five databases validate the effectiveness and robustness of our proposed SCCRC. Based on the experimental results on these databases, the following observations can be made:

(1) By enhancing the sparsity of representation of CRC, SCRC outperforms SRC and CRC in most cases, which reveals that sparseness of collaborative representation explicitly contributes to accurate classification of test samples.

(2) Thanks to the non-negative constraint on the coefficient vector, NRC achieves higher classification accuracy than SRC and CRC, which demonstrates the discriminative capability of the non-negative regularization term.

(3) By introducing the competitive representation term, CCRC is superior to CRC in terms of classification accuracy. This regularization term promotes competitive representation between distinct classes, which encourages the coefficient vector to be sparse to some extent.

(4) On the clean test images, improvement of CCRC-$\ell_1$ over CCRC is not that significant. However, on the corrupted test images, CCRC-$\ell_1$ outperforms CCRC by 10.14\%, which again verifies that sparsity of coefficient vector is necessary to improve the classification performance.

(5) Both on the clean and corrupted test images, our proposed SCCRC performs the best, which indicates that SCCRC needs clean training images. Like conventional RBCM, our proposed SCCRC is a general classification framework and it can be applied in other pattern classification tasks. For corrupted training images, as in \cite{iliadis2017robust}, we can first employ low rank matrix recovery techniques (e.g., robust PCA~\cite{candes2011robust} and its variants) to obtain clean training images, then SCCRC can be used for classification.}

\subsection{Parameter sensitiveness analysis}
\label{sec:4-8}
\begin{figure}[htbp]
  \centering
  \includegraphics[trim={0mm 0mm 0mm 0mm},clip, width = .6\textwidth]{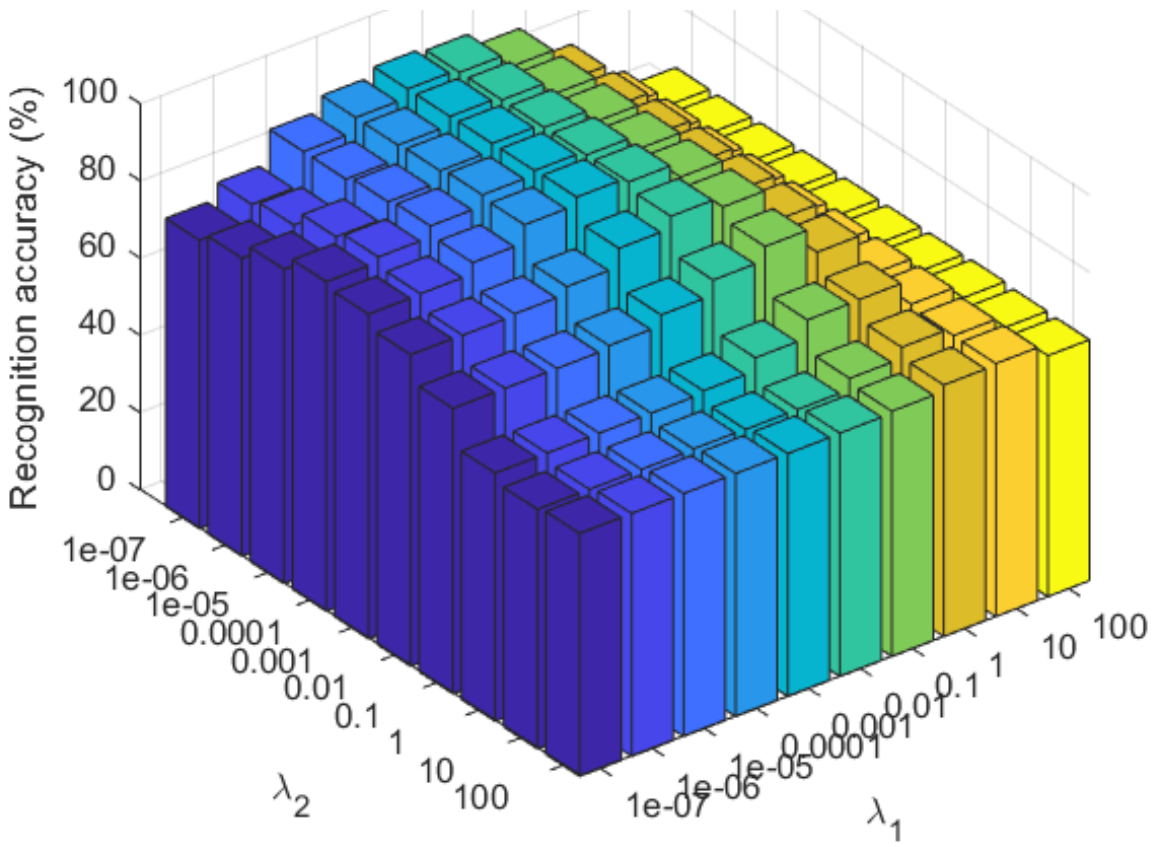}
  \caption{Recognition accuracy of SCCRC versus parameters $\lambda_{1}$ and $\lambda_{2}$ on the AR database.}
  \label{fig:bar3}
\end{figure}

\begin{figure}[htbp]
  \centering
  \includegraphics[trim={0mm 0mm 0mm 0mm},clip, width = .6\textwidth]{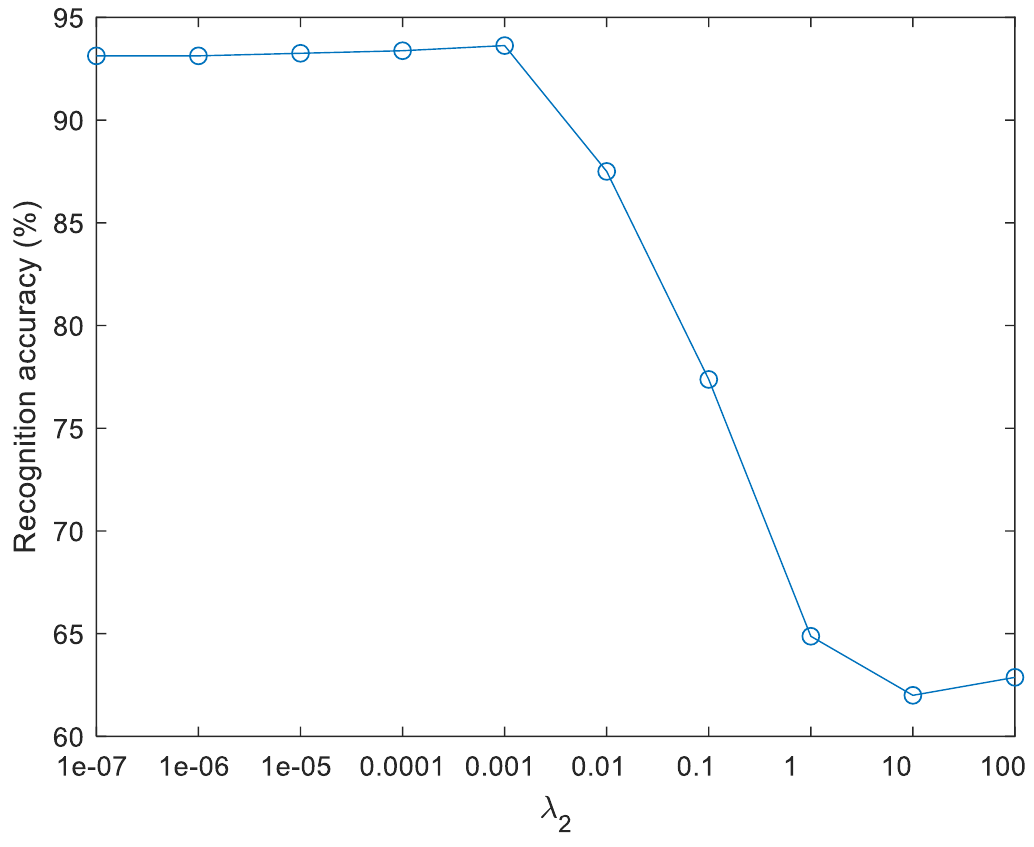}
  \caption{Recognition accuracy versus the variation of parameter $\lambda_{2}$ when $\lambda_{1}$=0.001.}
  \label{fig:lambda2}
\end{figure}

\begin{figure}[htbp]
  \centering
  \includegraphics[trim={0mm 0mm 0mm 0mm},clip, width = .6\textwidth]{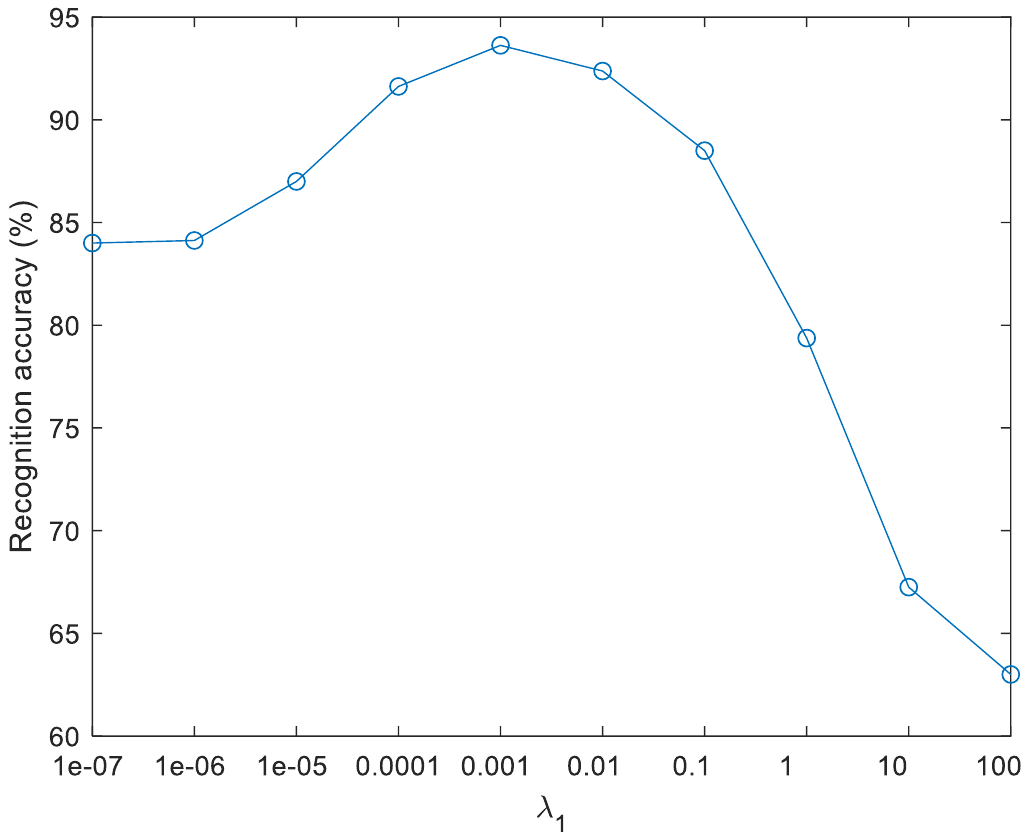}
  \caption{Recognition accuracy versus the variation of parameter $\lambda_{1}$ when $\lambda_{2}$=0.001.}
  \label{fig:lambda1}
\end{figure}

In our proposed SCCRC, there are three parameters to be determined, i.e. one parameter $\lambda$ for SRC and two parameters $\lambda_{1}$ and $\lambda_{2}$ for CCRC. We set $\lambda$=0.001 for SRC as in SCRC. To examine how the remaining parameters $\lambda_{1}$ and $\lambda_{2}$ influence the performance of SCCRC, we conduct experiments on the AR database. Experimental setting is the same as in Section \ref{sec:4-5} and the number of training samples per subject is 6. Fig. \ref{fig:bar3} illustrates the effect of parameter selection. We can see that SCCRC achieves superb results when both $\lambda_{1}$ and $\lambda_{2}$ are in proper range. More specifically, the performance is better when $\lambda_{2}$ is in the range of [0.0001 0.001], which validates the necessity of the competitive term in CCRC. Similarly, SCCRC can achieve better results when $\lambda_{1}$ is in the range of [0.0001 0.001]. To better illustrate the influence of $\lambda_{1}$ and $\lambda_{2}$, we plot the recognition accuracy against the variation of one parameter when the other is fixed. Fig. \ref{fig:lambda2} shows the influence of $\lambda_{2}$ when $\lambda_{1}$=0.001, we can see that the performance of SCCRC is desirable when $\lambda_{2}$ is assigned to a small value. Fig. \ref{fig:lambda1} presents the effect of $\lambda_{1}$ when $\lambda_{2}$=0.001, one can see that SCCRC performs stable when $\lambda_{1}$ is in the range of [0.0001 0.01]. Based on the above experimental results, we set $\lambda_{1}$=0.001 and $\lambda_{2}$=0.001 on the AR database.

\section{Conclusions}
\label{sec:sect_5}
Sparse representation based classification (SRC) and collaborative representation based classification (CRC) have been widely studied due to their promising results for classification. Although CRC reveals that it is the collaborative representation mechanism rather than the $\ell_{1}$-norm sparsity that makes SRC powerful, sparsity should not be completely ignored. Thus, in this paper, we propose a new technique to enhance the sparsity of CCRC by multiplying the coefficients of SRC and CCRC. Experiments on five publicly available databases validate the effectiveness of our proposed SCCRC, and demonstrate that SCCRC outperforms other representation based classification approaches.

In this paper, we did not explicitly consider the situation that both the training and test samples are contaminated due to occlusion or corruption, thus in future, we will extend SCCRC to tackle the above scenarios.

\begin{acknowledgements}
This work was supported in part by the National Natural Science Foundation of China (Projects Numbers: 61673194, 61672263, 61672265 and 61876072), and in part by the national first-class discipline program of Light Industry Technology and Engineering (Project Number: LITE2018-25).
\end{acknowledgements}

\bibliographystyle{spphys}       
\bibliography{mybibfile}   

\begin{thebibliography}{10}
\providecommand{\url}[1]{{#1}}
\providecommand{\urlprefix}{URL }
\expandafter\ifx\csname urlstyle\endcsname\relax
  \providecommand{\doi}[1]{DOI \discretionary{}{}{}#1}\else
  \providecommand{\doi}{DOI \discretionary{}{}{}\begingroup
  \urlstyle{rm}\Url}\fi

\bibitem{qu2018air}
X.~Qu, W.~Wang, K.~Lu, J.~Zhou, In-air handwritten chinese character
  recognition with locality-sensitive sparse representation toward optimized
  prototype classifier, Pattern Recognition \textbf{78}, 267 (2018)

\bibitem{prates2019kernel}
R.~Prates, W.R. Schwartz, Kernel cross-view collaborative representation based
  classification for person re-identification, Journal of Visual Communication
  and Image Representation \textbf{58}, 304 (2019)

\bibitem{yang2018hyperspectral}
J.~Yang, J.~Qian, Hyperspectral image classification via multiscale joint
  collaborative representation with locally adaptive dictionary, IEEE
  Geoscience and Remote Sensing Letters \textbf{15}(1), 112 (2018)

\bibitem{wright2009robust}
J.~Wright, A.Y. Yang, A.~Ganesh, S.S. Sastry, Y.~Ma, Robust face recognition
  via sparse representation, IEEE transactions on pattern analysis and machine
  intelligence \textbf{31}(2), 210 (2009)

\bibitem{li2010local}
C.G. Li, J.~Guo, H.G. Zhang, in \emph{2010 20th International Conference on
  Pattern Recognition} (IEEE, 2010), pp. 649--652

\bibitem{zhang2010k}
N.~Zhang, J.~Yang, in \emph{2010 Chinese conference on pattern recognition
  (CCPR)} (IEEE, 2010), pp. 1--5

\bibitem{ortiz2014face}
E.G. Ortiz, B.C. Becker, Face recognition for web-scale datasets, Computer
  Vision and Image Understanding \textbf{118}, 153 (2014)

\bibitem{aharon2006k}
M.~Aharon, M.~Elad, A.~Bruckstein, et~al., K-svd: An algorithm for designing
  overcomplete dictionaries for sparse representation, IEEE Transactions on
  signal processing \textbf{54}(11), 4311 (2006)

\bibitem{zhang2010discriminative}
Q.~Zhang, B.~Li, in \emph{2010 IEEE Computer Society Conference on Computer
  Vision and Pattern Recognition} (IEEE, 2010), pp. 2691--2698

\bibitem{jiang2013label}
Z.~Jiang, Z.~Lin, L.S. Davis, Label consistent k-svd: Learning a discriminative
  dictionary for recognition, IEEE transactions on pattern analysis and machine
  intelligence \textbf{35}(11), 2651 (2013)

\bibitem{kviatkovsky2017equivalence}
I.~Kviatkovsky, M.~Gabel, E.~Rivlin, I.~Shimshoni, On the equivalence of the
  lc-ksvd and the d-ksvd algorithms, IEEE transactions on pattern analysis and
  machine intelligence \textbf{39}(2), 411 (2017)

\bibitem{song2018euler}
Y.~Song, Y.~Liu, Q.~Gao, X.~Gao, F.~Nie, R.~Cui, Euler label consistent k-svd
  for image classification and action recognition, Neurocomputing \textbf{310},
  277 (2018)

\bibitem{zhang2011sparse}
L.~Zhang, M.~Yang, X.~Feng, in \emph{2011 International conference on computer
  vision} (IEEE, 2011), pp. 471--478

\bibitem{xu2017new}
Y.~Xu, Z.~Zhong, J.~Yang, J.~You, D.~Zhang, A new discriminative sparse
  representation method for robust face recognition via $ l\_ $\{$2$\}$ $
  regularization, IEEE transactions on neural networks and learning systems
  \textbf{28}(10), 2233 (2017)

\bibitem{zeng2017antinoise}
S.~Zeng, J.~Gou, L.~Deng, An antinoise sparse representation method for robust
  face recognition via joint l1 and l2 regularization, Expert Systems with
  Applications \textbf{82}, 1 (2017)

\bibitem{cai2016probabilistic}
S.~Cai, L.~Zhang, W.~Zuo, X.~Feng, in \emph{Proceedings of the IEEE conference
  on computer vision and pattern recognition} (2016), pp. 2950--2959

\bibitem{lan2018prior}
R.~Lan, Y.~Zhou, Z.~Liu, X.~Luo, Prior knowledge-based probabilistic
  collaborative representation for visual recognition, IEEE transactions on
  cybernetics  (2018)

\bibitem{yuan2018collaborative}
H.~Yuan, X.~Li, F.~Xu, Y.~Wang, L.L. Lai, Y.Y. Tang, A
  collaborative-competitive representation based classifier model,
  Neurocomputing \textbf{275}, 627 (2018)

\bibitem{zheng2019collaborative}
C.~Zheng, N.~Wang, Collaborative representation with k-nearest classes for
  classification, Pattern Recognition Letters \textbf{117}, 30 (2019)

\bibitem{waqas2013collaborative}
J.~Waqas, Z.~Yi, L.~Zhang, Collaborative neighbor representation based
  classification using l2-minimization approach, Pattern Recognition Letters
  \textbf{34}(2), 201 (2013)

\bibitem{gou2018new}
J.~Gou, L.~Wang, Z.~Yi, J.~Lv, Q.~Mao, Y.H. Yuan, A new discriminative
  collaborative neighbor representation method for robust face recognition,
  IEEE Access \textbf{6}, 74713 (2018)

\bibitem{akhtar2017efficient}
N.~Akhtar, F.~Shafait, A.~Mian, Efficient classification with sparsity
  augmented collaborative representation, Pattern Recognition \textbf{65}, 136
  (2017)

\bibitem{zeng2017multiplication}
S.~Zeng, X.~Yang, J.~Gou, Multiplication fusion of sparse and collaborative
  representation for robust face recognition, Multimedia Tools and Applications
  \textbf{76}(20), 20889 (2017)

\bibitem{li2016hyperspectral}
W.~Li, Q.~Du, F.~Zhang, W.~Hu, Hyperspectral image classification by fusing
  collaborative and sparse representations, IEEE Journal of Selected Topics in
  Applied Earth Observations and Remote Sensing \textbf{9}(9), 4178 (2016)

\bibitem{xu2019sparse}
J.~Xu, W.~An, L.~Zhang, D.~Zhang, Sparse, collaborative, or nonnegative
  representation: Which helps pattern classification?, Pattern Recognition
  \textbf{88}, 679 (2019)

\bibitem{naseem2010linear}
I.~Naseem, R.~Togneri, M.~Bennamoun, Linear regression for face recognition,
  IEEE transactions on pattern analysis and machine intelligence
  \textbf{32}(11), 2106 (2010)

\bibitem{deng2013defense}
W.~Deng, J.~Hu, J.~Guo, in \emph{Proceedings of the IEEE conference on computer
  vision and pattern recognition} (2013), pp. 399--406

\bibitem{deng2018face}
W.~Deng, J.~Hu, J.~Guo, Face recognition via collaborative representation: Its
  discriminant nature and superposed representation, IEEE transactions on
  pattern analysis and machine intelligence \textbf{40}(10), 2513 (2018)

\bibitem{combettes2005signal}
P.L. Combettes, V.R. Wajs, Signal recovery by proximal forward-backward
  splitting, Multiscale Modeling \& Simulation \textbf{4}(4), 1168 (2005)

\bibitem{samaria1994parameterisation}
F.S. Samaria, A.C. Harter, in \emph{Proceedings of 1994 IEEE Workshop on
  Applications of Computer Vision} (IEEE, 1994), pp. 138--142

\bibitem{goel2005face}
N.~Goel, G.~Bebis, A.~Nefian, in \emph{Biometric Technology for Human
  Identification II}, vol. 5779 (International Society for Optics and
  Photonics, 2005), vol. 5779, pp. 426--438

\bibitem{phillips1997feret}
P.J. Phillips, H.~Moon, P.~Rauss, S.A. Rizvi, in \emph{Proceedings of IEEE
  Computer Society Conference on Computer Vision and Pattern Recognition}
  (IEEE, 1997), pp. 137--143

\bibitem{georghiades2001few}
A.S. Georghiades, P.N. Belhumeur, D.J. Kriegman, From few to many: Illumination
  cone models for face recognition under variable lighting and pose, IEEE
  Transactions on Pattern Analysis \& Machine Intelligence (6), 643 (2001)

\bibitem{martinez2007ar}
A.~Mart{\'\i}nez, R.~Benavente, The ar face database, Computer Vision Center,
  Technical Report \textbf{24} (1998)

\bibitem{yang2010fast}
A.Y. Yang, S.S. Sastry, A.~Ganesh, Y.~Ma, in \emph{2010 IEEE International
  Conference on Image Processing} (IEEE, 2010), pp. 1849--1852

\bibitem{wen2018low}
J.~Wen, X.~Fang, Y.~Xu, C.~Tian, L.~Fei, Low-rank representation with adaptive
  graph regularization, Neural Networks \textbf{108}, 83 (2018)

\bibitem{li2015learning}
S.~Li, Y.~Fu, Learning robust and discriminative subspace with low-rank
  constraints, IEEE transactions on neural networks and learning systems
  \textbf{27}(11), 2160 (2015)

\bibitem{iliadis2017robust}
M.~Iliadis, H.~Wang, R.~Molina, A.K. Katsaggelos, Robust and low-rank
  representation for fast face identification with occlusions, IEEE
  Transactions on Image Processing \textbf{26}(5), 2203 (2017)

\bibitem{candes2011robust}
E.J. Cand{\`e}s, X.~Li, Y.~Ma, J.~Wright, Robust principal component analysis?,
  Journal of the ACM (JACM) \textbf{58}(3), 11 (2011)

\end{thebibliography}

\end{document}